\documentclass[a4paper, 10pt, conference]{ieeeconf}      

\IEEEoverridecommandlockouts                              

\usepackage{amsmath} 
\usepackage{amssymb}  
\usepackage{pdfpages}
\usepackage{siunitx}
\usepackage{textcomp}
\usepackage{float}
\usepackage{url}

\title{\LARGE \bf
Optimal Trajectory Generation for Autonomous Vehicles Under Centripetal Acceleration Constraints for In-lane Driving Scenarios
}

\author{Yajia Zhang*, Hongyi Sun, Jinyun Zhou, Jiangtao Hu, Jinghao Miao 
\thanks{Yajia Zhang, Hongyi Sun, Jinyun Zhou, Jiangtao Hu, Jinghao Miao are with Baidu USA LLC,
        {250 Caribbean Drive Sunnyvale, USA}  *Corresponding author: Yajia Zhang 
        {\tt\small zhangyajia@baidu.com} \newline
        \textbf{The paper was accepted by IEEE International Conference on Intelligent Transportation Systems (ITSC) 2019}}%
}

\begin{document}

\maketitle
\thispagestyle{empty}
\pagestyle{empty}

\begin{abstract}
This paper presents a noval method that generates optimal trajectories for autonomous vehicles for in-lane driving scenarios. The method computes a trajectory using a two-phase optimization procedure. In the first phase, the optimization procedure generates a close-form driving guide line with differetiable curvatures. In the second phase, the procedure takes the driving guide line as input, and outputs dynamically feasible, jerk and time optimal trajectories for vehicles driving along the guide line. This method is especially useful for generating trajectories at curvy road where the vehicles need to apply frequent accelerations and decelerations to accommodate centripetal acceleration limits.
\end{abstract}

\section{Introduction}
Trajectory planning is an important component in autonomous driving systems (ADS). It plays a critical role on safety and comfort. Safety is of top priority as any collision might lead to hazardous situations. Assuming predicted trajectories of surrounding obstacles are given from upper stream module of ADS, path-time obstacle graph is a commonly used tool for collision avoidance analysis if future path of the autonomous driving vehicle (ADV) is determined. This method projects the predicted trajectories of surrounding obstacles onto the spatio-time plane and forms path-time obstacles which specify at which time the further path of the ADV would be on collision. The free area forms the collision-free zone for trajectory planning. This method is particularly useful for autonomous vehicles in structured road scenarios. It fully utilizes the domain knowledge, as most vehicles are driving along lanes. The method we propose adopts path-time-obstacle graph in collision avoidance analysis and always plans a trajectory that lie within the collision-free zone.

Comfort is another goal to achieve for ADS. Several factors affect and are used to measure the comfort of one trajectory. Acceleration and acceleration change rate (commonly known as jerk) are most commonly used metrics for vehicle trajectories. Furthermore, depending on the direction, human weight acceleration and jerk significant differently for longitudinal and lateral movement. Acceleration and jerk in lateral direction must be bounded and minimized. For driving along a curvy road, the longitudinal speed must be adjusted frequently according to the curve, i.e., the curvature of the road. A driving guide line is an abstraction of the road center line, which contains the geometrical information of the road. We assume the target of the autonomous vehicle in-lane driving is following the driving guide line. To achieve comfortable riding experience, the vehicle needs to accelerate and decelerate according to the curvature of the driving guide line. In our proposed method, the algorithm can directly consider the geometrical information of the driving guide line. 

Optimization is a common approach in trajectory generation as it takes the objective or cost function and constraints directly into trajectory generation. For high degrees of freedom (DOFs) configuration space, optimization for trajectory generation is generally slow and prone to local minima, it is generally suitable for lower dimensional vehicle configuration space. In our method, we use a two-phase optimization procedure. Each one intends to solve a subset of trajectory generation problem. In this way, it greatly reduces the overall complexity of optimization. For the first phase, our method generates a smooth driving guide line for ADV to follow; in the second phase, the optimization procedure takes the collision-free zone resulted from path-time obstacle graph analysis and the close-formed driving guide line as input, and generates a collision-free and comfort trajectory that minimizes longitudinal acceleration, and centripetal acceleration and jerk.




\section{Related Work}
Trajectory planning is a critical component in autonomous driving systems. Recently, a number of algorithms \cite{5940562, miller2008team, urmson2008autonomous} have been developed since DARPA Grand Challenge (2004, 2005) and Urban Challenge (2007). 

Randomized planners such as Rapidly Exploring Random Tree (RRT)\cite{lavalle2001randomized} are intended to solve high-DOF robot motion planning with differential constraints. However, it is difficult for randomized planners to utilize the domain knowledge from the structured environment for quickly convergence. Nevertheless, the computed trajectory is generally low quality and thus cannot be used directly without a post-processing step. Recent research on optimal randomized planner, such as \cite{hwan2011anytime}, can produce high-quality trajectories given enough planning time. But the convergence to optimal trajectory takes rather long time thus it cannot be used in the dynamically changing environment.

Discrete search method \cite{kuwata2009real} computes a trajectory by concatenating a sequence of pre-computed maneuvers. The contatenation is done by checking whether the ending state of a maneuver is sufficiently close to the starting state of the target maneuver. This method generally works well for simple environment such as highway scenarios. However, the number of required maneuvers needs to grow exponentially in order to solve complex urban driving cases.


The work in \cite{ziegler2014making} runs an quadratic programming procedure in global/map frame. The trajectory is finely discretized in Cartesian space. The positional attributions of the trajectory are directly used as optimization variables, and outputs a trajectory that minimizes the objective function which combines the measurement of safety and comfort. The advantage of using optimization is it provides direct enforcement of optimality modeling. The dense discretization approach provides maximal control of trajectory to tackle complex scenarios.


In \cite{werling2010optimal}, trajectory planning is performed in Frenet frame. Given a smooth driving guide line, this method decouples the movement of vehicle in map frame into two orthogonal movements, one longitudinal movement that along the driving guide line and one lateral movement that perpendicular to the guide line. For both movements, the trajectories are generated using random samples, i.e., the end conditions and parameter are discretized into certain resolution. The sampled end conditions are directly connected with the initial condition using quintic or quartic polynomials. Then these longitudinal and lateral trajectories are combined and selected according to a predefined cost function. 

The major drawback with the method is it lacks control of the trajectory. For complex driving scenarios, it is difficult for this method to generate feasible trajectories. Nevertheless, some caveats of using polynomial include problems with stopping, unexpected acceleration and deceleration. To tackle complex problems in real world, we need to maximize our ability of controlling the trajectory. Thus, optimization is a promising direction as constraints in the task domain can be directly considered in trajectory generation. Existing optimization-based methods such as \cite{ziegler2014making} \cite{schlechtriemen2016wiggling} runs the optimization in map frame. 

The overall algorithm framework we proposed is similar to the one in \cite{werling2010optimal}, however, we use optimization to solve the 1d planning problems, which greatly enhances the flexibility of the trajectory.

The method we present hybrids the Frenet frame trajectory planning framework and optimization-based trajectory generation. Also, the trajectory optimization is performed in two-phases, where each phase is a lower dimensional problem. In this proposed method, the difficulty of optimization is greatly reduced.

\section{Problem Definition}
The configuration for a vehicle with differential constraints in Cartesian space can be represented using three variables, $(x, y, \theta)$, where $x$, $y$ specify the coordinate of some reference point for the vehicle and $\theta$ specifies the vehicle's heading angle in the Cartesian space. In our work, we incorporate one more dimension $\kappa$, which is the instant curvature resulted from vehicle's steering, into the configuration space for more accurate configuration modeling, and hence the computed trajectory provides additional information that can be used for better designing the feedback controller. Trajectory planning for non-holonomic vehicles is essentially finding a function ${\boldsymbol{\tau}}(t)$ that maps a time $t$ to a specific configuration $(x, y, \theta, \kappa)$.

\subsection*{Trajectory Planning in Frenet frame}
\begin{figure}
    \centering
    \includegraphics[width=0.41\textwidth]{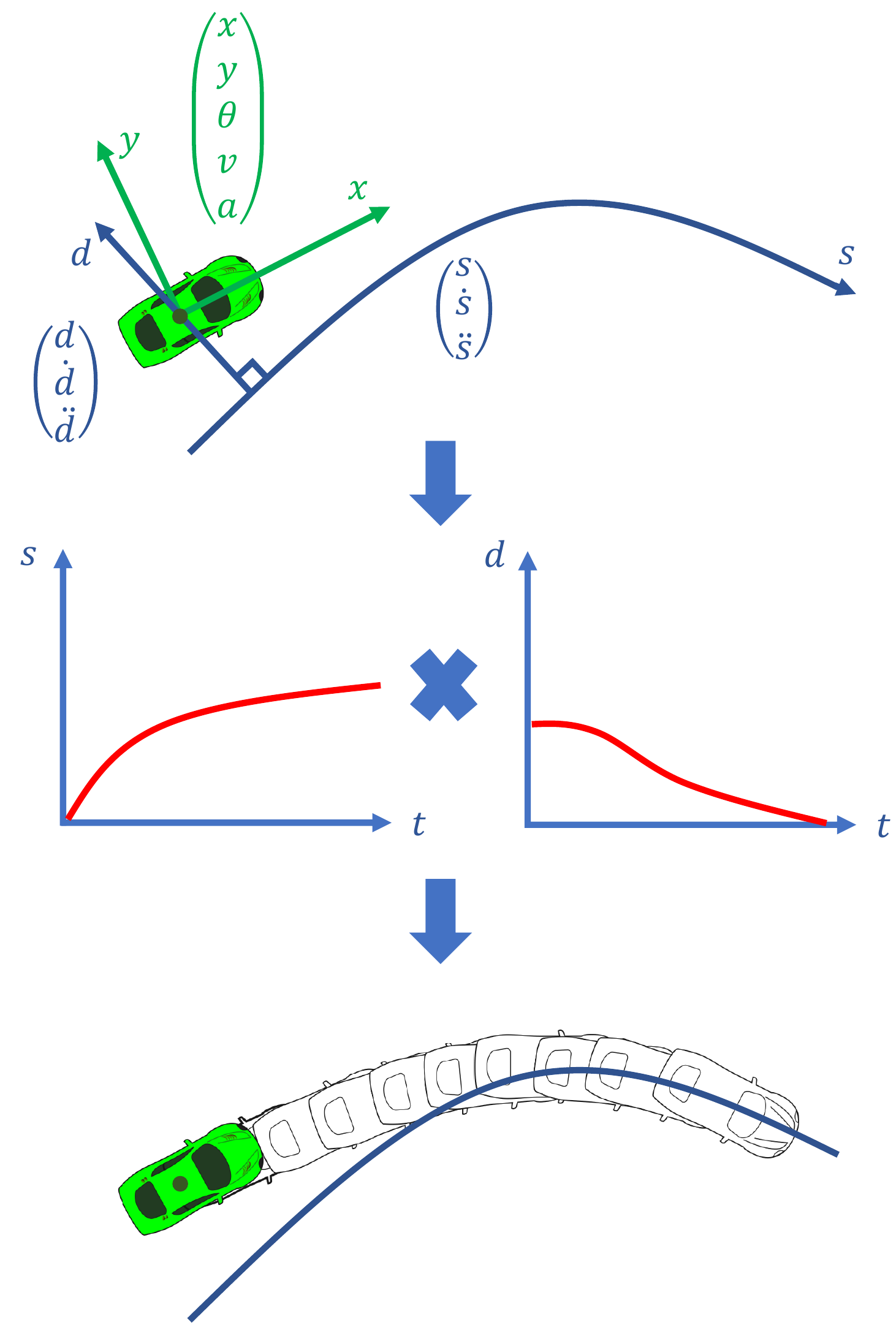}
    \caption{Illustration of vehicle trajectory planning in assist of Frenet frame. First, the vehicle dynamic state $(x, y, \theta, \kappa, v, a)$, which represents vehicles position, heading, steering angle, velocity and acceleration, respective, is projected on to a given driving guide line to obtain its decoupled states in Frenet frame. $(s, \dot{s}, \ddot{s})$ represents the vehicle state, i.e., position, velocity and acceleration, along the guide line (i.e., longitudinal state) and $(d, \dot{d}, \ddot{d})$ represents the vehicle state, i.e., position, velocity and acceleration, perpendicular to the guild line (i.e., lateral state). Then, plan longitudinal and lateral motions \textbf{independently}. Finally, longitudinal and lateral motions in Frenet frame are combined and transformed to a trajectory in Cartesian space.}
    \label{fig:frenet-frame-framework}
\end{figure}

Our method adopts a similar framework as in \cite{werling2010optimal}, which utilizes the concept of Frenet frame for trajectory planning (see Fig. \ref{fig:frenet-frame-framework}). Given a smooth driving guide line, a Frenet frame decouples the vehicle motions in Cartesian space into two independent 1D movements, longitudinal movement that moves along the guide line and lateral movement that moves orthogonally to the guide line. Thus, a trajectory planning problem in Cartesian space is transformed to two lower dimensional and independent planning problems in Frenet frame. This framework exploits the task domain that most vehicles are moving along the lane, and it is particularly advantageous as it greatly simplifies problem by reducing the dimensionality of planning. 

We assume that the autonomous vehicle is roughly driving around the guild line. The main task for in lane driving is to generate a trajectory for the autonomous vehicle driving along the guide line.
In this paper, we assume the vehicle will stay close to the guide line and possible lateral deviations to the guide line can be corrected by the controller. Thus, lateral state of the vehicle is always zero. The trajectory planning problem in our work is now reduced to finding a function $\boldsymbol{s}(t)$, where $s$ is the longitudinal coordinate along the guide line. For any given time $t$, $\boldsymbol{s}(t)$ returns a vehicle longitudinal state $(s, \dot{s}, \ddot{s})$ at time $t$. $\boldsymbol{s}(t)$  along with a zero function $\boldsymbol{d}(t)$ forms a trajectory in Cartesian space. 

\section{Phase I: Smooth Driving Guide Line Generation}
A guide line is the prerequisite for planning in Frenet frame. The smoothness of a guide line is critical to generating high quality trajectories as the vehicle's velocity, acceleration, heading and steering information are implicitly encoded in the guide line. According to Cartesian-Frenet frame conversions (see \cite{werling2010optimal} for details), to obtain continuous acceleration and steering angles, the derivative of curvature of the guide line must be continuous as well.

Generally, the guide line is obtained from map in the form of a sequence of coordinates in map frame, i.e, $(x_0, y_0), \ldots, (x_{n-1}, y_{n-1})$, without having the necessary geometrical information, such as curve tangent angle . Before the guide line can be used, a smoothing phase that assigns the necessary geometrical information is needed. In our implementation, we use a non-linear optimization procedure for guide line smoothing as optimization provides direct control of optimality and constraint satisfactions. 

To formulate an optimization for guide line smoothing, we consider the following aspects:

\begin{enumerate}
    \item \textbf{Objective} Minimal wiggling of geometrical properties. The wiggling of the geometrical properties will lead to unsmooth acceleration and wiggling of vehicle steering. The objective of the optimization is designed to minimize the length of the guide line $s$, curvature $\kappa$ and curvature derivative $\dot{\kappa}$ (i.e., $d{\kappa}/ds$).
    \item \textbf{Line Representation} The representation of driving guide line is flexible enough for complex road shapes. We choose a discrete-continuous hybridyzation for the guide line representation: the input points in Cartesian space work as "knots" and close-form curves connect consecutive input points with continuous geometrical properties at the joint points. For a raw guide line with $n$ input points, $n - 1$ pieces of curves are used.
    \item \textbf{Constraints} The resulted guide line do not necessarily need to pass through input points as the Cartesian coordinates might have errors from map generation; however, the deviations to input points must be limited within certain threshold to preserve the original shape. This requirement is formulated as constraints in optimization.
\end{enumerate}




In our implementation, we are inspired by \cite{gulati2013nonlinear} to use polynomial spiral curve as piecewise baseis. Spiral curve, which represents its shape using a function of curve tangent angle $\theta$ w.r.t. accumulated curve length $s$, is the key element in guide line smoothing. The advantage of using spiral curves is the geometric properties, such as curve direction $\theta$, curvature $\kappa$ ($d\theta / ds$) and curvature change rate $\dot{\kappa}$ ($d\theta^2 / ds^2$) can be easily derived by just taking the derivative of the function $\boldsymbol{\theta}(s)$ w.r.t. $s$ directly. Compared to other curve formulations, such as parametric curves in Cartesian space, the objective function can be greatly simplified.




In the guide line smoothing problem, we use quintic polynomial spiral curves to connect consecutive input points. The following are the variables in the optimization where $\theta_i$, $\dot{\theta}_i$ and $\ddot{\theta}$ represent curve tangent angle, curvature and curvature change rate for some input point $i$ and $\Delta s_i$ is the length of the spiral curve that connects point $i$ and $i+1$:

\begin{figure}[!htb]
    \centering
    \includegraphics[width=0.45\textwidth]{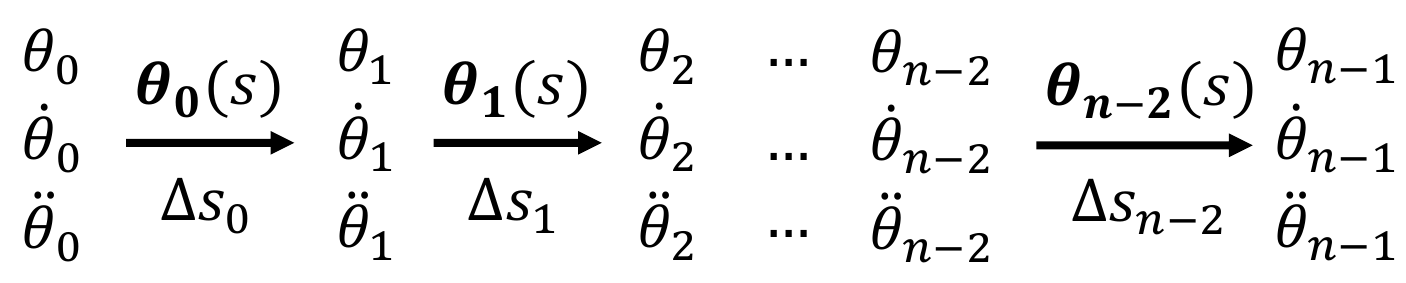}
\end{figure}


The geometrical property variables $\theta_i$, $\dot{\theta}_i$, $\ddot{\theta}_i$, $\theta_{i+1}$, $\dot{\theta}_{i+1}$, $\ddot{\theta}_{i+1}$ and the curve length variable $\Delta s_i$ uniquely determines the coefficients $c_j$ of a quintic polynomial:
\begin{align*}
    \boldsymbol{\theta}_i(s) = \sum_{j=0}^{5} c_j * s^j
\end{align*}

The objective function for optimization is approximated by evaluating a predefined $m$ equal spacing internal points for each piecewise spiral curve. The optimization procedure finds the optimal values for variables $\theta$, $\dot{\theta}$, $\ddot{\theta}$ and $\Delta s$ that minimizes the objective function. The result of guide line optimization is a sequence of quintic spiral curves that are smoothly connected at the input points. By taking the derivative w.r.t. $s$, we are able to obtain a sequence of smooth quartic polynomials that represent curvature of the guide line, where curvature and curvature derivative info are smoothly connected at the input points.

\begin{align*}
    \boldsymbol{\kappa}_i(s) = \sum_{j=1}^{5} j * c_j * s^{j-1}
\end{align*}

Commonly, the guide line is finely discretized to certain resolution for trajectory generation. However, keeping the piecewise \textbf{close-form of curvature representation} is the key to Phase II.


\newcommand\ftr{15}
\newcommand\ftrside{15}

\begin{figure*}[t]
\label{fig:uturn}
\centering
\begin{tabular}{ccc}
\label{fig:u_turn_guide_line}
      \includegraphics[trim={\ftrside} {\ftr} {\ftrside} {\ftr}, width=0.3\textwidth]{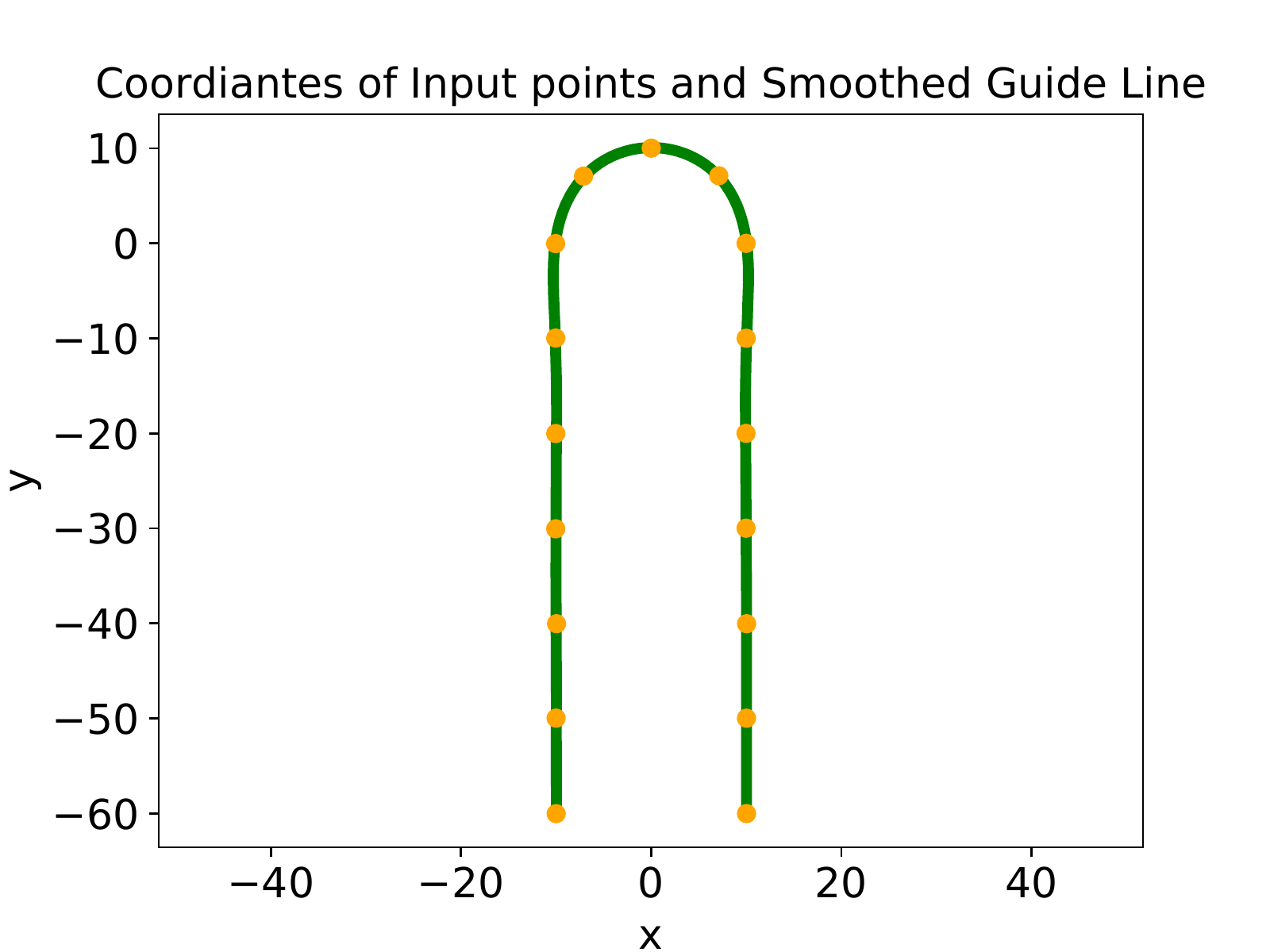} & 
      \includegraphics[trim={\ftrside} {\ftr} {\ftrside} {\ftr}, width=0.3\textwidth]{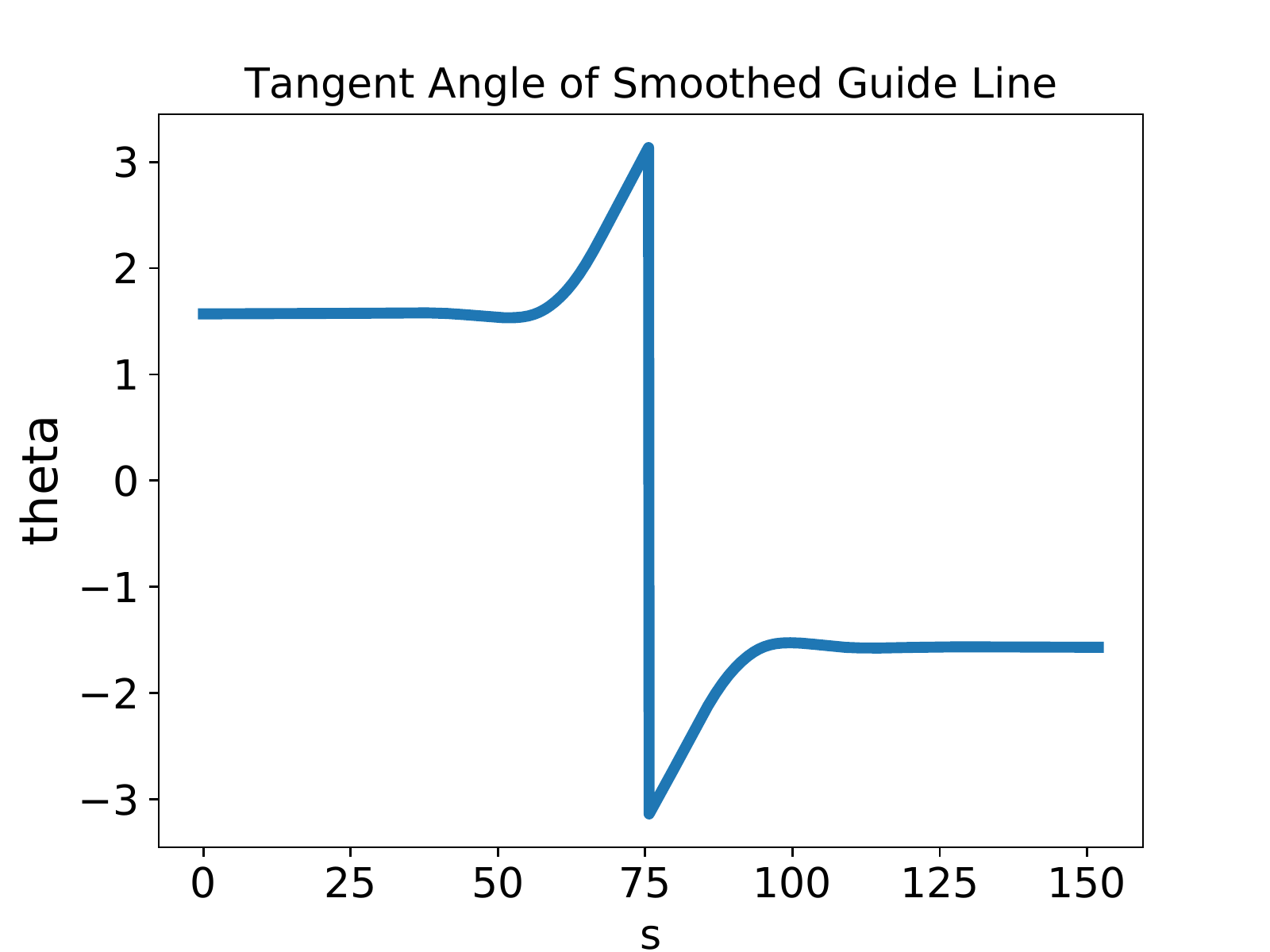}  &
      \includegraphics[trim={\ftrside} {\ftr} {\ftrside} {\ftr}, width=0.3\textwidth]{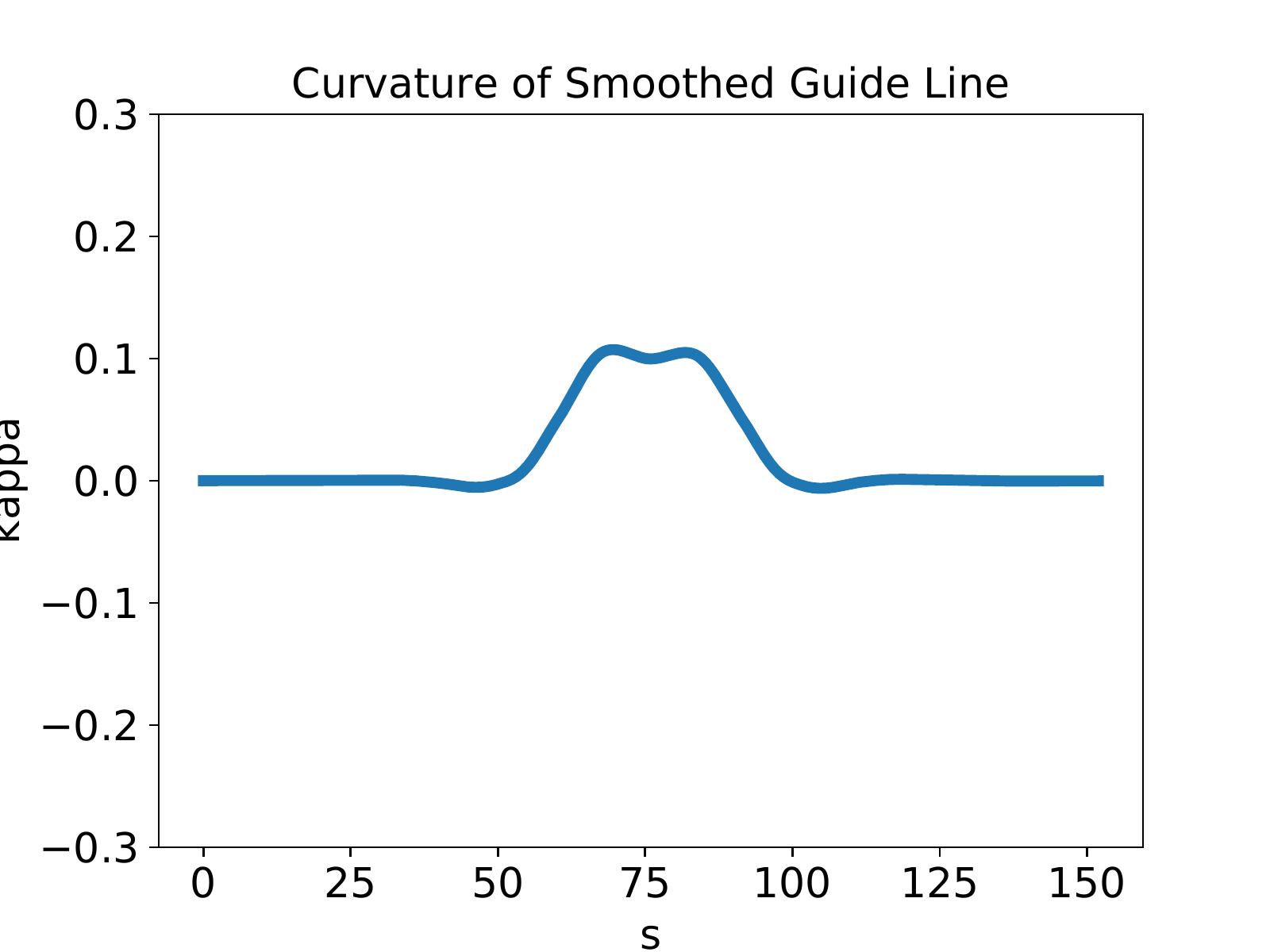} \\
\end{tabular}
\caption{An example of guide line smoothing. The first figure shows the guide line in Cartesian space, which describes a common U-turn scenario with turning radius 10 $m$. The input is a sequence of discretized points (in yellow) and the output (in green) is a sequence of spiral paths that connect the input points. The input points are deliberately perturbed with random noise within 0.1 $m$. The second figure shows the tangent direction of the optimized guide line and the third figure shows the curvature.}
\end{figure*}

\section{Phase II: Trajectory Generation}
\subsection{Optimality Modeling}
\label{subsec:optimality_modeling}
The goal of trajectory generation for autonomous vehicles is to compute a trajectory that drives the passengers to the destination safely and comfortably. The following are key factors that are commonly used to measure the quality of a trajectory:

\subsubsection{\textbf{Task achievement}} The most important application of autonomous driving is letting the vehicle transport the passenger to a predefined destination. For in-lane driving scenarios, basically there are two tasks: 
\begin{itemize}
    \item \textbf{Cruise} The vehicle reaches a target cruise speed. When a front obstacle with lower speed exists, the vehicle maintains safe distance to the front obstacle.
    \item \textbf{Stop} The vehicle stops at a certain point. When the vehicle reaches a predefined destination, or an intermediate target, e.g., a red light or stop sign, etc, it proceeds with a full stop.
\end{itemize}

The task achievement can be evaluated by measuring the difference between task specific target state and vehicle's state at the end of trajectory. 

\subsubsection{\textbf{Safety}}
\begin{figure}
    \centering
    \includegraphics[width=0.35\textwidth]{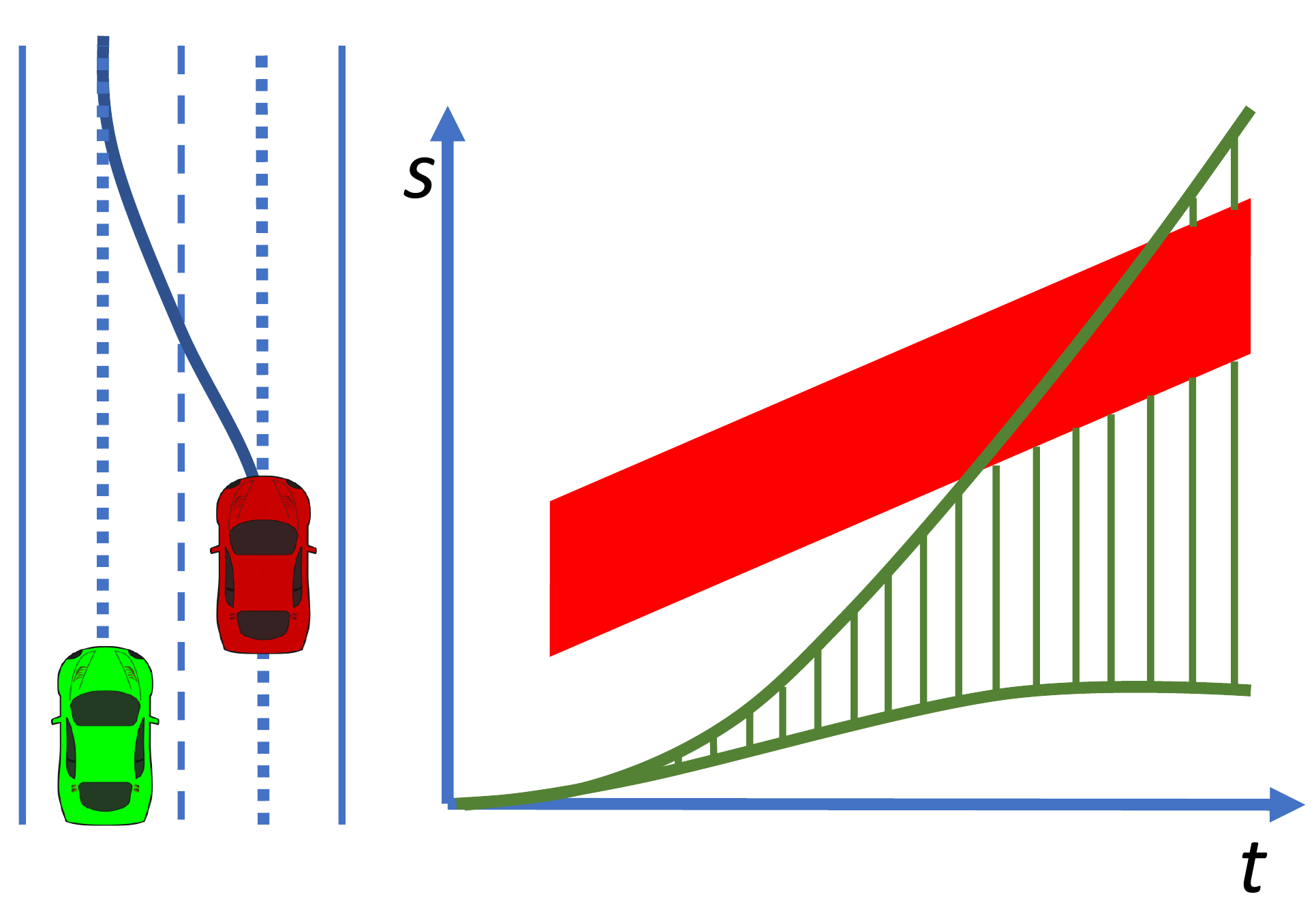}
    \caption{Path-time-obstacle graph for one driving scenario. Left: Scenario illustration. Ego vehicle (in green) drives along the center of the lane while obstacle vehicle (in red) on adjacent lane tries to change to the left lane. Right: Path-time-obstacle graph. The s axis represents the accumulated length of the center lane of ego vehicle; the t axis represents time. The path-time-obstacle (approximated) is shown in red which it represents at certain time t, which part of the center lane would be blocked by the obstacle, considering the geometry of the obstacle vehicle. Two quadratic lines represents the boundaries of reachable portions on the center lane by considering the dynamical limits, e.g., velocity and acceleration bounds, of the ego vehicle.}
    \label{fig:path-time-graph}
\end{figure}

We limit the application of proposed algorithm for in-lane driving scenarios, i.e., driving along a predetermined guide line. To avoid collisions with obstacles in the environment, we use the concept of \textbf{path-time-obstacle} graph. The path-time decomposition was firstly introduced by Kant and Zucker \cite{kant1986toward} to tackle the problem of dynamic obstacle avoidance along a given path under velocity constraints. We assume the path (driving guide line) for ego vehicle is pre-determined, and trajectories for nearby obstacles can be accurately predicted. The essential idea of path-time-obstacle graph is computing which time $t$ that which portion $s$ of the predetermined guide line is blocked by obstacles (see Fig. \ref{fig:path-time-graph}). 
The output of the collision avoidance analysis is allowable region(s) $[s_{min}, s_{max}]$ for a given time $t$. If the ego vehicle at time $t$ stays within the allowable regions, collision to obstacles is avoided. In complex cases, it is possible that the allowable regions for a given time $t$ are not unique. In our work, simple heuristics is used to uniquely determine a region by considering the dynamics of the vehicle, however, a formal situation reasoning module with more advanced techniques can surely provide better decisions. The result of reasoning can be abstracted as a function $\boldsymbol{s}_{free}(t)$ that returns a region $(s_{min}, s_{max})$ given any time $t$. A safety margin can also be added by increasing $s_{min}$ or decreasing $s_{max}$ for the autonomous driving vehicle to keep certain distance to front obstacle.


\subsubsection{\textbf{Comfort}}
Comfort is an important metric of trajectory quality. In our formulation, the comfort is modeled using the following factors:
\begin{itemize}
\item Longitudinal acceleration: minimization of longitudinal acceleration to reduce frequent throttle or brake input.
\item Longitudinal acceleration change rate (jerk): minimization of jerk in longitudinal direction, i.e., minimization of input changes of throttle or brakes. 
\item Centripetal acceleration: minimization of centripetal acceleration along the trajectory. When the vehicle approaches a curve, it must decelerate gracefully to reduce the centripetal acceleration resulted from the curve.
\item Time: for accomplishing certain task, e.g., reaching a target velocity, the time is minimized.
\end{itemize}

\subsection{Constraints}
As our algorithm is intended for vehicle driving in structured environment, we assume the vehicle is always moving forward monotonically. Based on vehicle's dynamical properties, the following constraints must be satisfied at any time $t$ of the trajectory with time range $t_\tau$ to achieve safety and comfort: 

\begin{itemize}
    \item Longitudinal velocity bounds \\ $\boldsymbol{\dot{s}}(t) \in [0, \dot{s}_{max}]$
    \item Longitudinal acceleration bounds \\ $\boldsymbol{\ddot{s}}(t) \in [\ddot{s}_{min}, \ddot{s}_{max}]$
    \item Longitudinal jerk bounds \\ $\boldsymbol{\dddot{s}}(t) \in [\dddot{s}_{min}, \dddot{s}_{max}]$
    \item Centripetal acceleration bounds \\ $\boldsymbol{a}_{\boldsymbol{c}}(t) \in [-\dot{a}_{c_{max}}, \dot{a}_{c_{max}}]$
\end{itemize}

Besides dynamical constraints, to achieve collision avoidance, the spatial value $s$ for any given time $t$ must lie within the allowable regions given by $\boldsymbol{s}_{free}(t)$:

\begin{itemize}
    \item Safety bounds \\
    $\boldsymbol{s}(t) \in \boldsymbol{s}_{free}(t)$
\end{itemize}

\subsection{Trajectory Formulation and Discretization}
\label{subsec:formulation}
To effectively compute an optimal trajectory while evaluating constraint satisfaction in practice, a common approach is discretizing the trajectory $\boldsymbol{s}(t)$ according to a certain parameter. The objective function is then approximated by evaluating these discretized points, and constraint satisfaction is checked on these points as well. 

Depending on the parameter we use for trajectory discretization, the formulation of optimization, including objective function, constraint formulation, can be quite differently. In this section, we discuss two choices of parameters for discretization and their pros and cons. Despite difference on choice of parameters for discretization, both methods discretize the trajectory to second-order derivative of $s$, i.e., $\ddot{s}$ to achieve accurate controlling to the acceleration level, and assume a constant third-order term $\dddot{s}$ (i.e., jerk) connects consecutive discretized points. 

\subsubsection{Spatial Parameter Discretization}
\label{subsubsec:spatial_space_discretization}
The first choice is discretizing the trajectory using spatial parameter $s$ to a predefined resolution $\Delta s$. The resulting trajectory is represented by the following variables, assuming the starting point with $s = 0$:

\begin{figure}[!htb]
    \centering
    \includegraphics[width=0.40\textwidth]{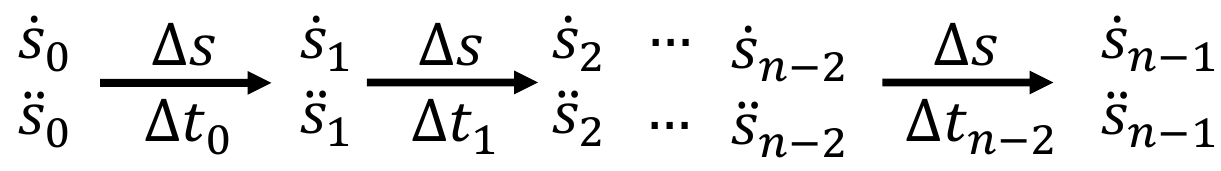}
\end{figure}


In this formulation, between consecutive discretized point is an equal spatial distance $\Delta s$. The optimization variables include the velocity $\dot{s}$, acceleration $\ddot{s}$ at each point and the time intervals $\Delta t$ between consecutive points. As we have discussed in previous section, the curvature of the guide line $\boldsymbol{\kappa}$ is a function of spatial parameter $s$. 

The major advantage of spatial parameter discretization is the spatial value $s$ for each discretized point is fixed, thus the curvature. Therefore, the speed upper limit resulted from the centripetal acceleration constraint can be precomputed and remains fixed throughout the optimization. However, there are several drawbacks with this formulation:


\begin{itemize}
    \item Difficult objective function for optimization. For one term in optimality modeling, longitudinal jerk minimization, the jerk is computed using the differencing between consecutive points. In this formulation, there would be a division term between optimization variables in objective function, which leads to a difficult non-convex optimization problem.
    \item Difficult to enforce safety bound constraint. For path-time-obstacle analysis, the output is a function $\boldsymbol{s}_{free}(t)$ that maps a time $t$ to a safe region $(s_{min}, s_{max})$. In this formulation, as the time $t$ is a variable, the safe region for one point changes as time $t$ changes.
\end{itemize}

Because of these issues, despite the formulation of centripetal acceleration constraint is simplified, it is not a good choice to discretize the trajectory using spatial parameter.

\subsubsection{Temporal Parameter Discretization}
\label{subsubsec:temporal_discretization}
In the contrast, we propose temporal parameter discretization, which discretizes the trajectory using time $t$ to a given resolution $\Delta t$. The resulting trajectory is represented by the following variables, assuming the starting point with $t = 0$:

\begin{figure}[!htb]
    \centering
    \includegraphics[width=0.45\textwidth]{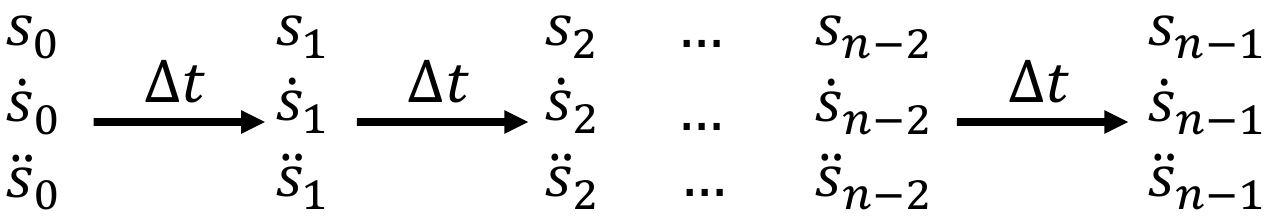}
\end{figure}

In this formulation, the optimization variables include the spatial parameter $s$, velocity $\dot{s}$ and acceleration $\ddot{s}$ at each discretized point. Compared to spatial-space discretization, this formulation avoids the divisional problem between variables for computing the longitudinal jerk term. Also, it is easier to incorporate safety bound constraint as safety bound as the safety bound remains fixed for each point throughout the optimization. 

A major challenge with temporal parameter discretization is the difficulty to consider the constraints and objectives involving centripetal acceleration, in contrast to spatial parameter discretization. Centripetal acceleration at one point is decided by two factors: the velocity $\dot{s}$, and the curvature $\kappa$. Since $\kappa$ is associated with spatial parameter $s$, as $s$ changes iteratively in the optimization procedure, the curvature $\kappa$ changes correspondingly. In other words, the velocity limit for one discretized point is now a function involving $\boldsymbol{\kappa}$(s). Assuming the maximal allowed centripetal acceleration is a fixed value ${a_c}_{max}$, the constraint for centripetal acceleration limit is as follows:

\begin{align*}
    -{a_c}_{max} \leq \dot{s}^2 * \boldsymbol{\kappa}(s) \leq {a_c}_{max}
\end{align*}

In order to apply the constraint to an optimization procedure, the term $\dot{s}^2 * \boldsymbol{\kappa}(s)$ must be differentiable. One of our major contributions is using a sequence of close-form spiral paths directly as the representation of guide line and thus making the term differentiable. 

Another difficulty of temporal parameter discretization is modeling the preference of minimal time for task accomplishment. To model the preference, we introduce a reference velocity term in the objective formulation. The reference velocity models the preferred velocity without considering the obstacle and curves. The integral of difference between current velocity and reference velocity is included in the objective function. This term encourages the vehicle to reach the reference velocity with minimal time spent. In next section, we discuss the detailed optimization formulation using temporal parameter discretization.

\subsection{Complete Optimization Formulation}
Following the discussion at \ref{subsec:optimality_modeling}, suppose a trajectory $\boldsymbol{s}(t)$ is parameterized by time $t$ with maximal time length $t_{\tau}]$, the objective function is designed as follows:

\begin{align}
    \boldsymbol{f} &= w_{\ddot{s}} * \int_{0}^{t_{\tau}} \boldsymbol{\ddot{s}}(t)^2 dt \\ 
               &+ w_{\dddot{s}} * \int_{0}^{t_{\tau}} \boldsymbol{\dddot{s}}(t)^2 dt \\ 
               &+ \, w_{a_c} * \int_{0}^{t_{\tau}} \boldsymbol{\dot{s}}(t)^2 * \boldsymbol{\kappa}(\boldsymbol{s}(t)) dt \\ 
               &+\, w_{\dot{s}_{ref}} * \int_{0}^{t_{\tau}} (\boldsymbol{\dot{s}}(t) - \dot{s}_{ref})^2 dt \\
               &+ \, w_{s_{task}} * (\boldsymbol{s}(t_{\tau}) - s_{task})^2 \\
               &+ \, w_{\dot{s}_{task}} * (\boldsymbol{\dot{s}}(t_{\tau}) - \dot{s}_{task})^2 \\
               &+ \, w_{\ddot{s}_{task}} * (\boldsymbol{\ddot{s}}(t_{\tau}) - \ddot{s}_{task})^2 
\end{align}

where $\dot{s}$, $\ddot{s}$ are the linear velocity and acceleration respectively. $\boldsymbol{\kappa}$ is a function that takes a spatial parameter $s$ as input and outputs the curvature $\kappa$ at the given point. Term 1-3 directly model the longitudinal acceleration, jerk and centripetal acceleration respectively. Term 4 implicitly models the minimization of time for accomplishing certain task. This term encourages the trajectory quickly reach a reference speed $\dot{s}_{ref}$. Term 5-7 are used to model the task preference, which encourage the vehicle to reach certain predefined state by the end of trajectory.

As we discretize the trajectory using temporal parameter, the objective function can be approximated by evaluating the $n$ discretized points. 

\begin{align*}
    \boldsymbol{\tilde{f}} &= w_{\ddot{s}} * \sum_{i=0}^{n-1} \ddot{s}_i^2 \\ 
              &+ \, w_{\dddot{s}} * \sum_{i=0}^{n-2} \dddot{s}_{i \rightarrow i + 1}^2 \\ 
              &+ \, w_{a_c} * \sum_{i=0}^{n-1} \dot{s}_i^2 * \boldsymbol{\kappa}(s_i)  \\ 
              &+ \, w_{\dot{s}_{ref}} * \sum_{i=0}^{n-1}(\dot{s}_i - \dot{s}_{ref})^2 \\
              &+ \, w_{s_{task}} * (s_{n-1} - s_{task})^2 \\
              &+ \, w_{\dot{s}_{task}} * (\dot{s}_{n-1} - \dot{s}_{task})^2 \\
              &+ \, w_{\ddot{s}_{task}} * (\ddot{s}_{n-1} - \ddot{s}_{task})^2
\end{align*}

As discussed in \ref{subsec:formulation}, a constant jerk term $\dddot{s}$ is used to connect consecutive points. Its value can be computed by differencing the acceleration terms:

\begin{align*}
    \dddot{s}_{i \rightarrow i + 1} = \frac{\ddot{s}_{i+1} - \ddot{s}_{i}}{\Delta t}
\end{align*}

Constraint checking is performed on these discretized points as well: $\forall i \in \mathbb{Z}, i \in [0, n-1]$,

\begin{align*}
    s_i & \in \boldsymbol{s}_{free}(i*\Delta t) \\
    \dot{s}_i &\in [0, \dot{s}_{max}]  \\
    \ddot{s}_i &\in [\ddot{s}_{min}, \ddot{s}_{max}] \\
    \dddot{s}_{i \rightarrow i + 1} & \in [\dddot{s}_{min}, \dddot{s}_{max}] \\
    a_{c_i} &\in [-a_{c_{max}}, a_{c_{max}}], \, a_{c_i} = \dot{s}_i^2 * \boldsymbol{\kappa}(s_i)
\end{align*}



Nevertheless, to maintain the continuity of the piecewise constant-jerk trajectory, the following equality equations must be satisfied between consecutive points:

\begin{align*}
    \ddot{s}_{i+1} &= \ddot{s}_{i} + \int_{0}^{\Delta t} \dddot{s}_{i} dt \\
                   &= \ddot{s}_{i} +  \dddot{s}_{i} * \Delta{t}         \\
    \dot{s}_{i+1} &= \dot{s}_{i} + \int_{0}^{\Delta t} \ddot{s}_{i+1} dt \\
                  &= \dot{s}_{i} + \ddot{s}_{i} * \Delta{t} + \frac{1}{2} *  \dddot{s}_i * \Delta{t}^2 \\
    s_{i+1} &= s_i + \int_{0}^{\Delta t} \dot{s}_i dt \\
            &= s_i + \dot{s}_{i} * \Delta{t} + \frac{1}{2} * \ddot{s}_{i} * \Delta{t}^2 + \frac{1}{6} * \dddot{s}_{i \rightarrow i + 1} * \Delta{t}^3 
\end{align*}

\newcommand\w{0.225}
\newcommand\tr{135}
\newcommand\trside{200}

\begin{figure*}[t]
\label{fig:traj}
\centering
\begin{tabular}{cccc}
\label{fig:u_turn_traj}
      \includegraphics[trim={\trside} {\tr} {\trside} {\tr}, width=\w\textwidth]{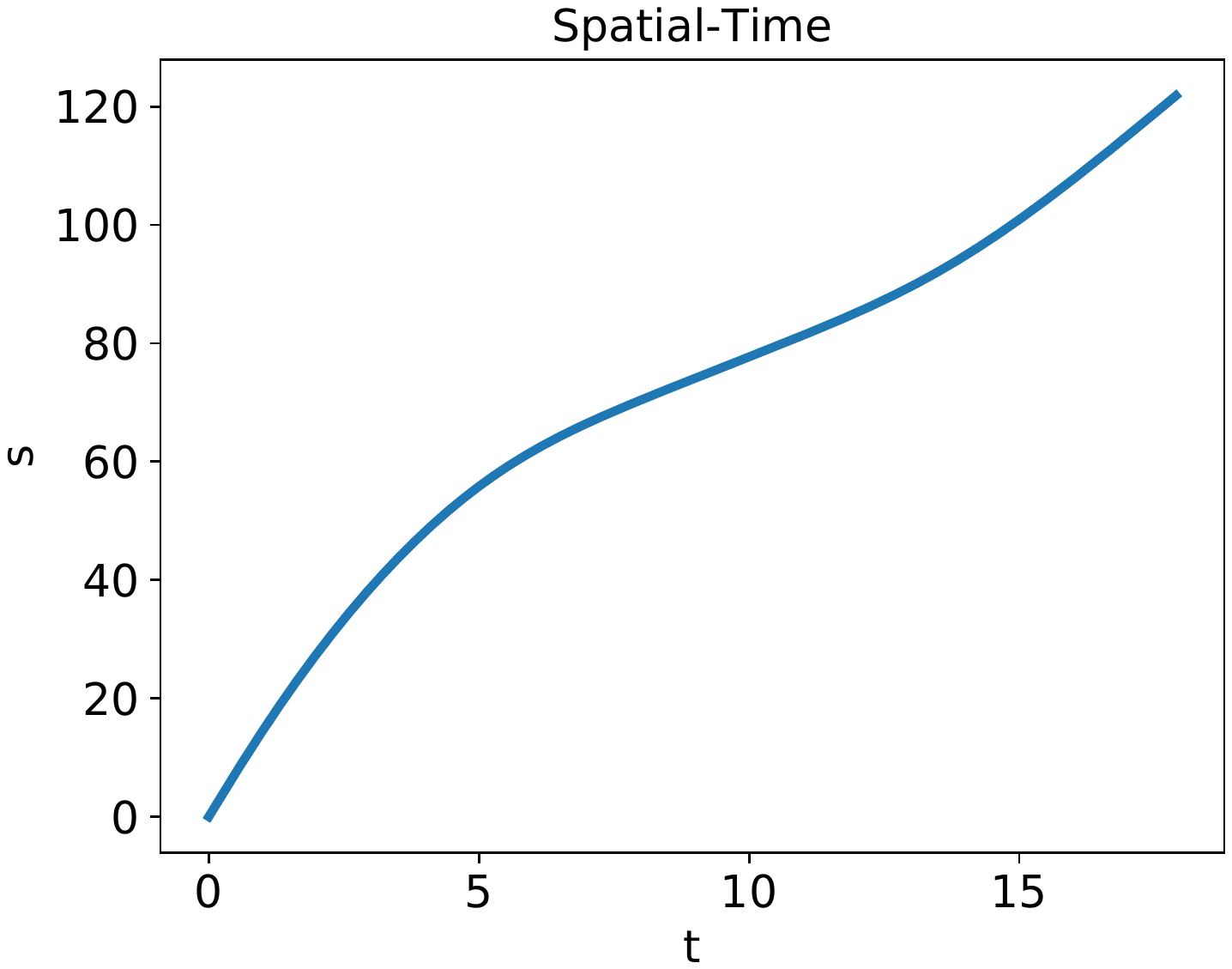} & 
      \includegraphics[trim={\trside} {\tr} {\trside} {\tr}, width=\w\textwidth]{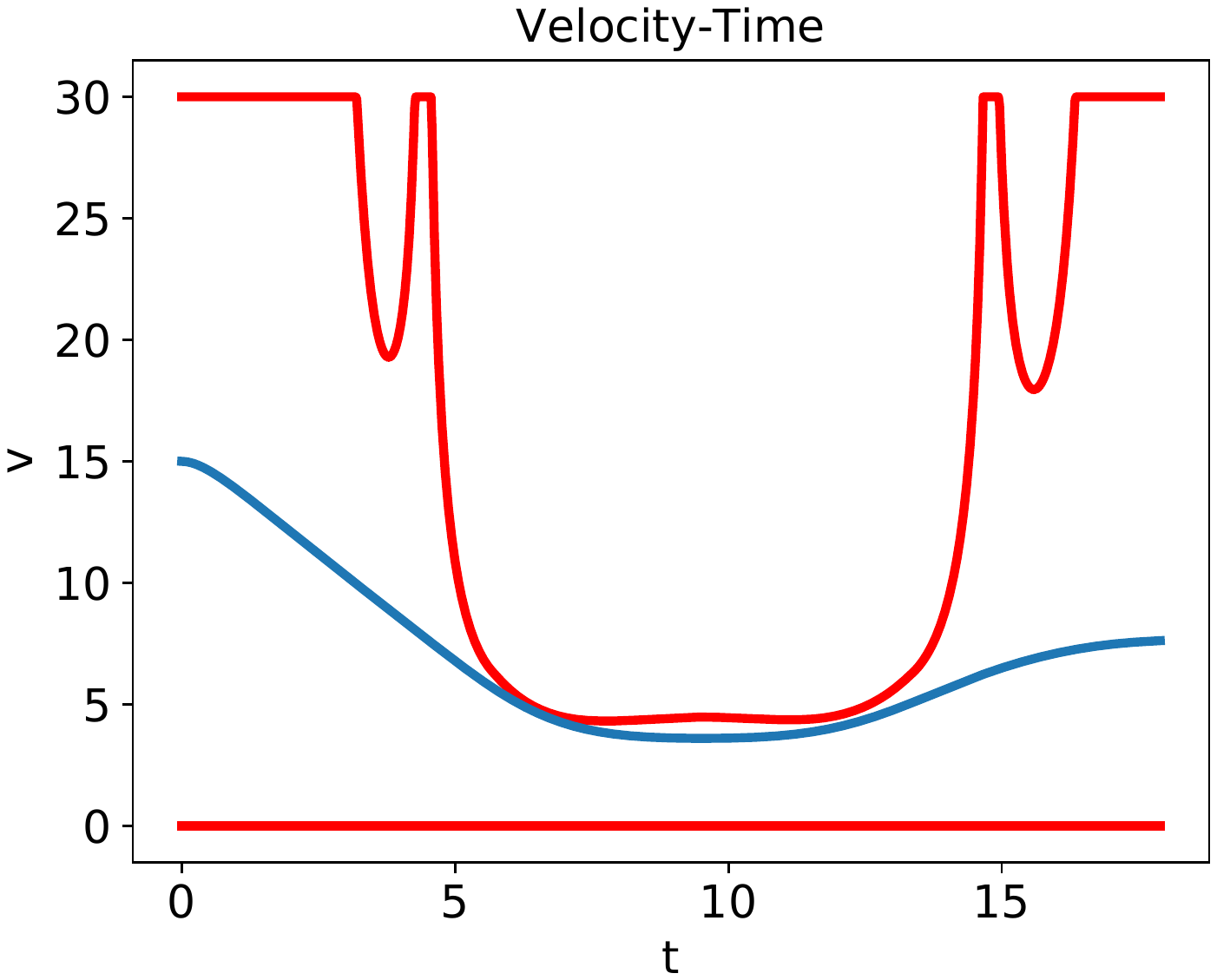} &
      \includegraphics[trim={\trside} {\tr} {\trside} {\tr}, width=\w\textwidth]{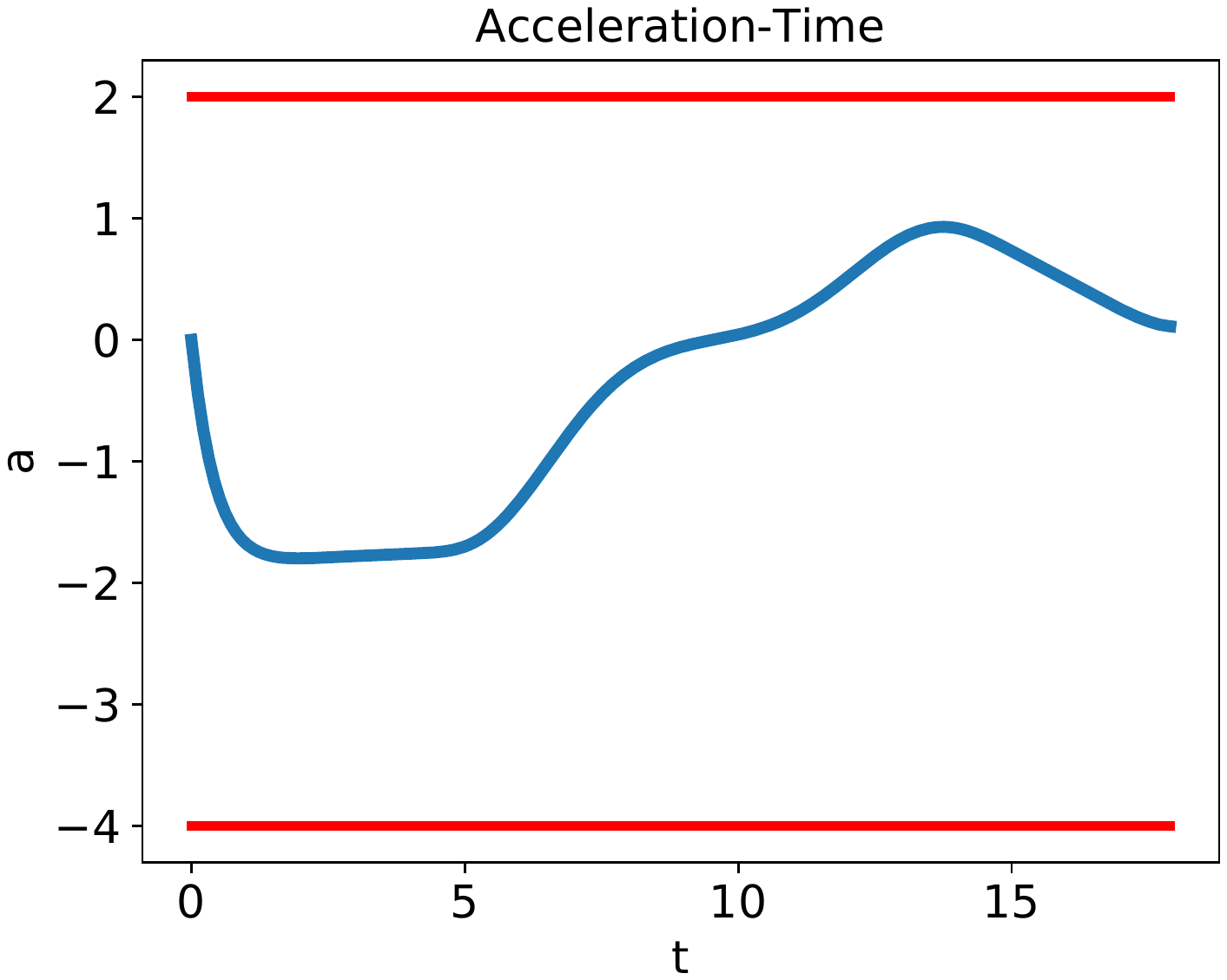} &
      \includegraphics[trim={\trside} {\tr} {\trside} {\tr}, width=\w\textwidth]{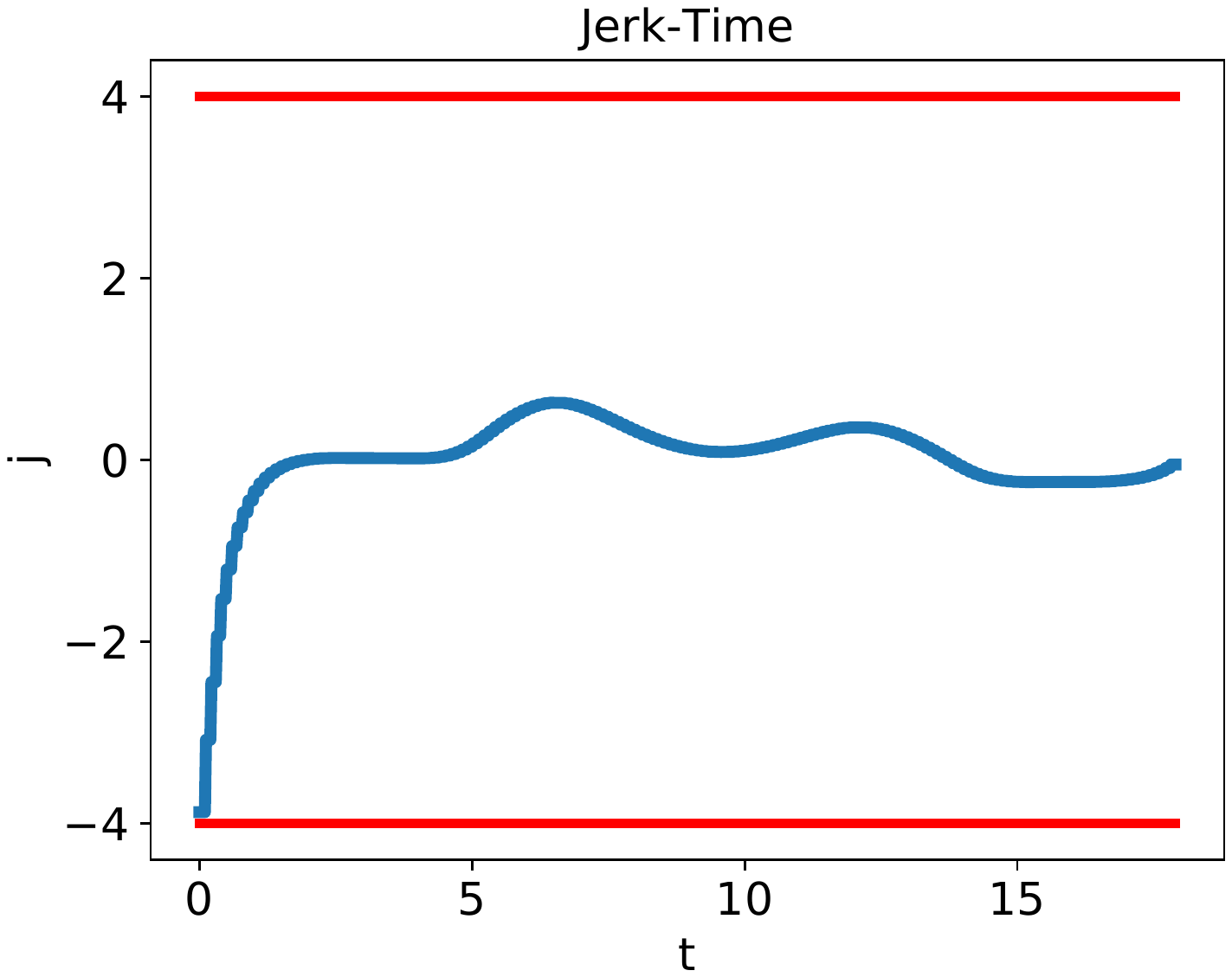} \\
      \includegraphics[trim={\trside} {\tr} {\trside} {\tr}, width=\w\textwidth]{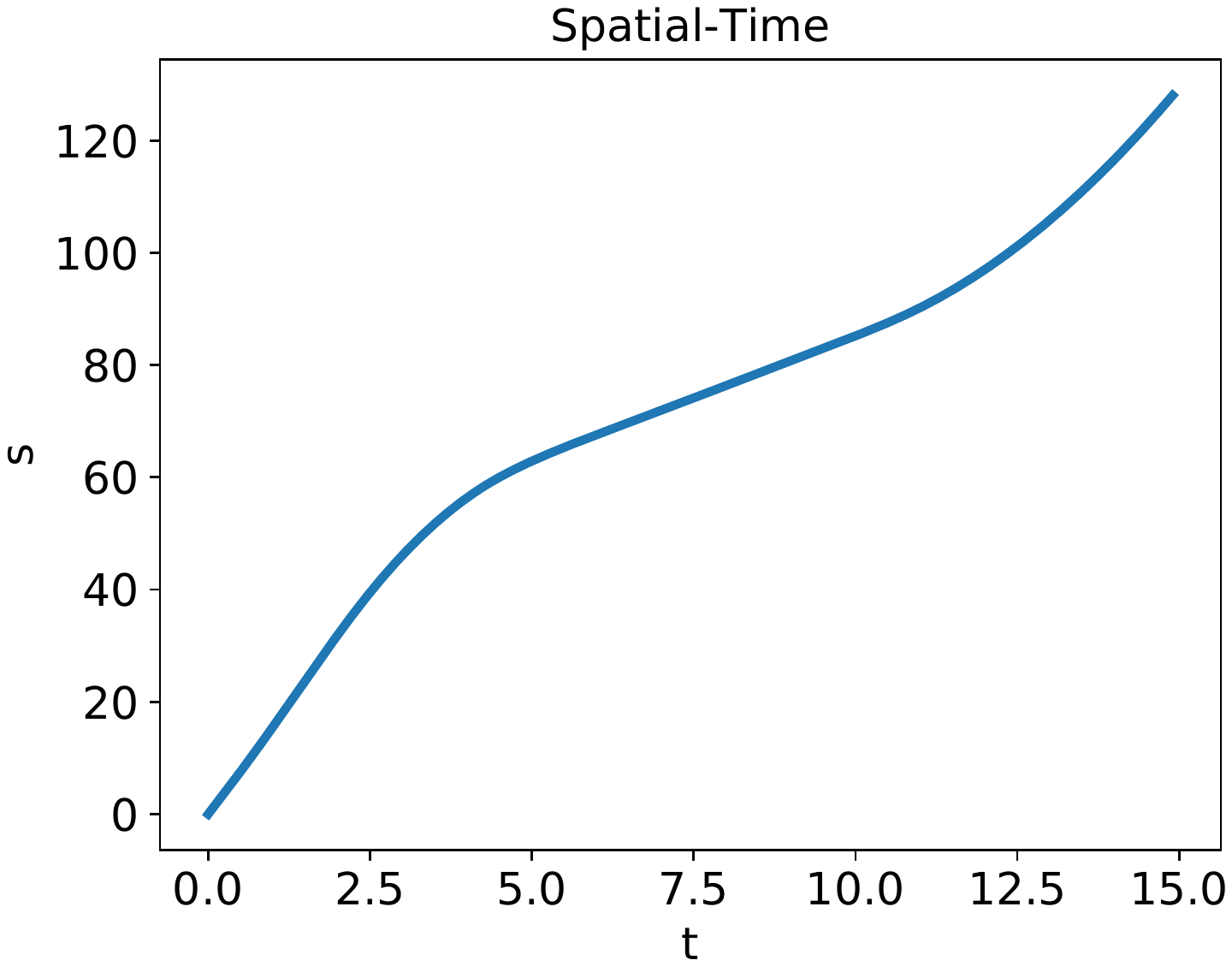} & 
      \includegraphics[trim={\trside} {\tr} {\trside} {\tr}, width=\w\textwidth]{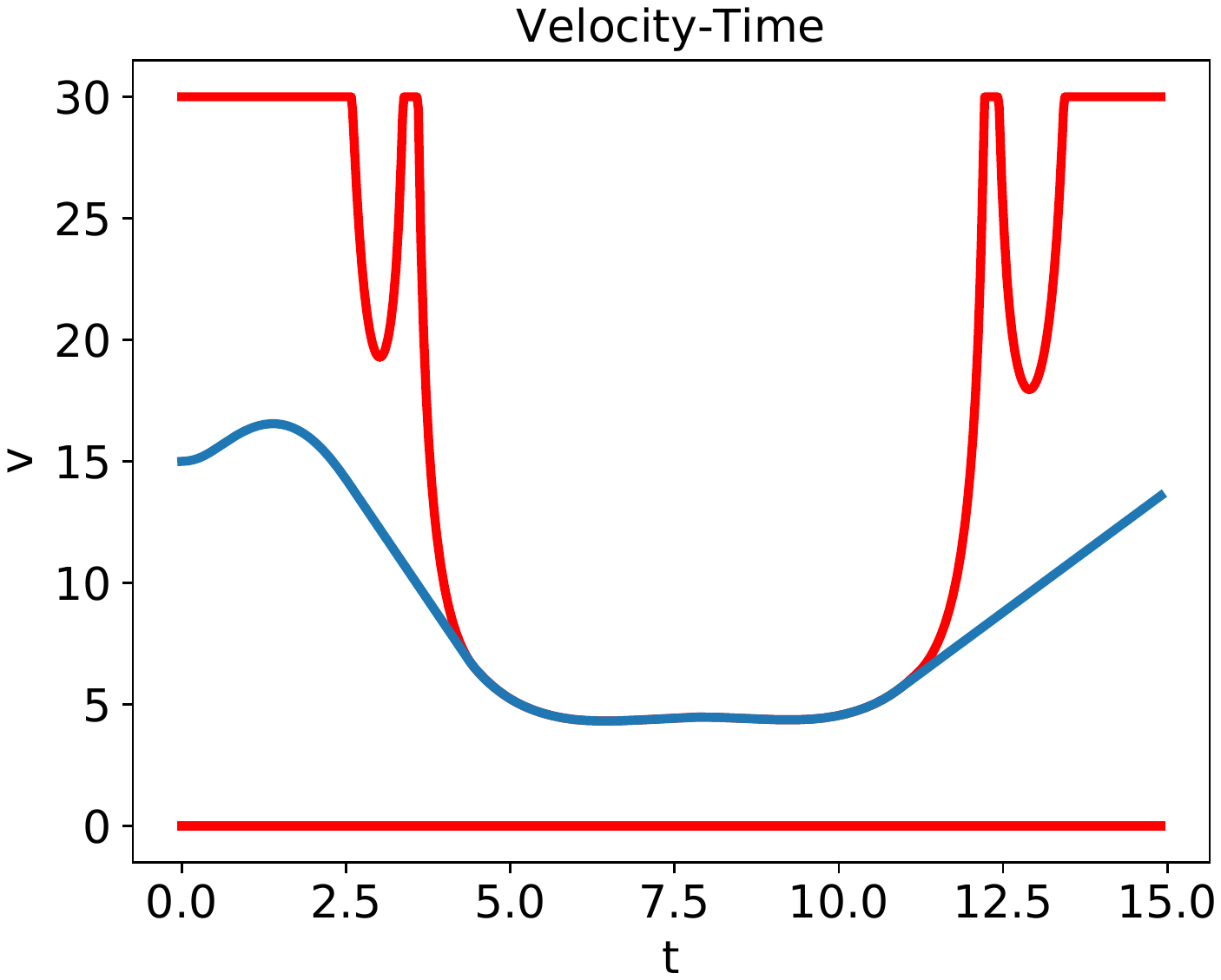} &
      \includegraphics[trim={\trside} {\tr} {\trside} {\tr}, width=\w\textwidth]{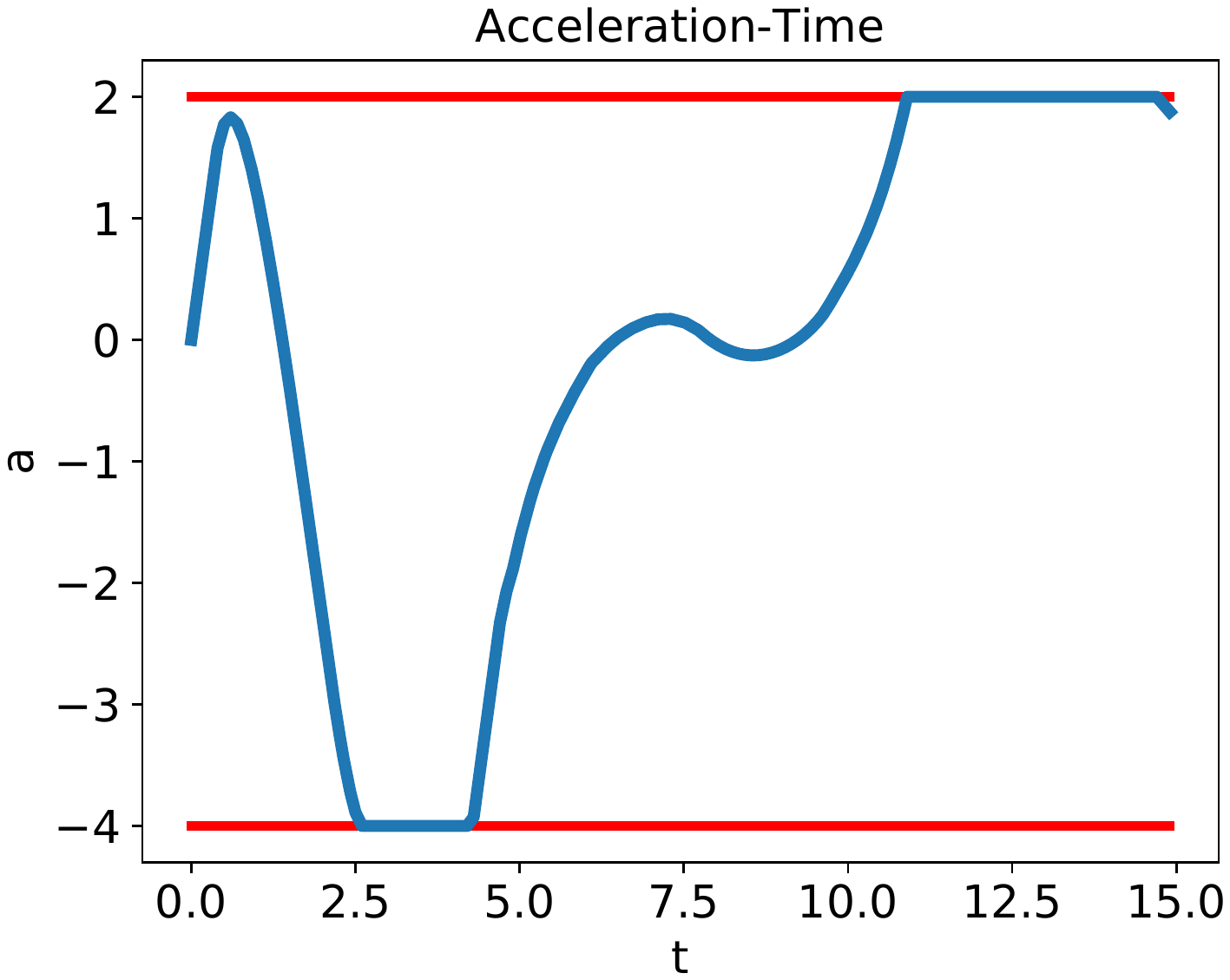} &
      \includegraphics[trim={\trside} {\tr} {\trside} {\tr}, width=\w\textwidth]{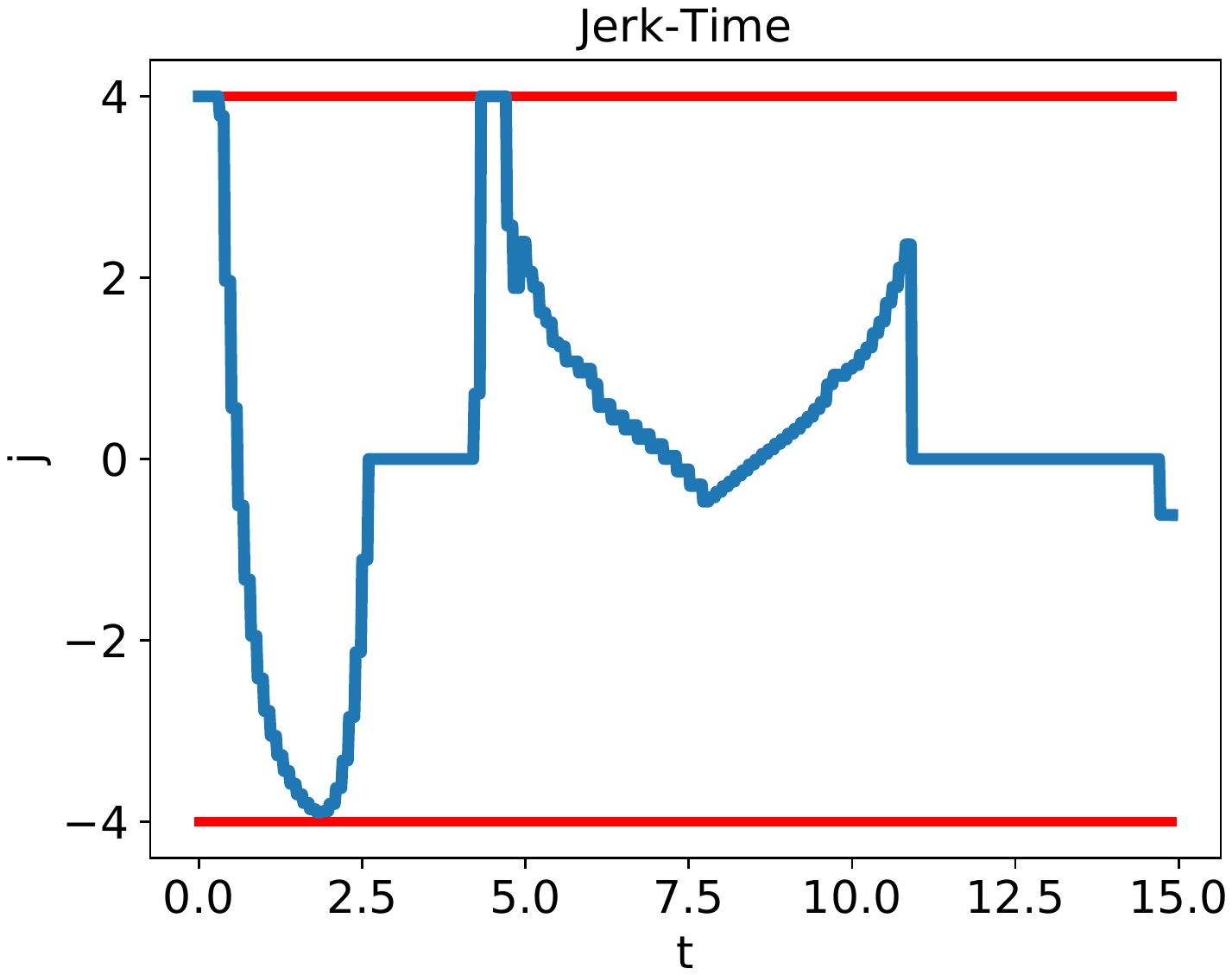} \\
      \includegraphics[trim={\trside} {\tr} {\trside} {\tr}, width=\w\textwidth]{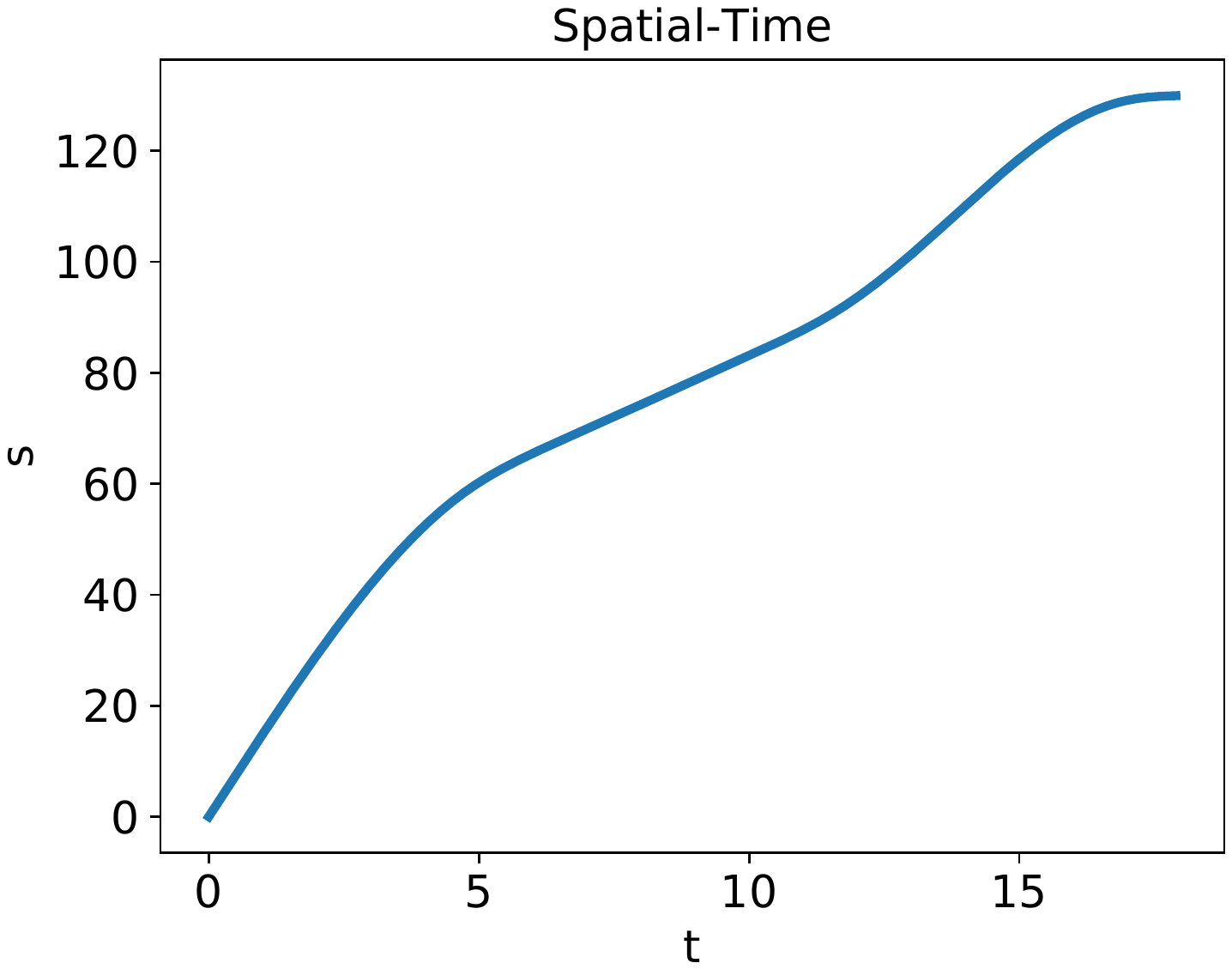} & 
      \includegraphics[trim={\trside} {\tr} {\trside} {\tr}, width=\w\textwidth]{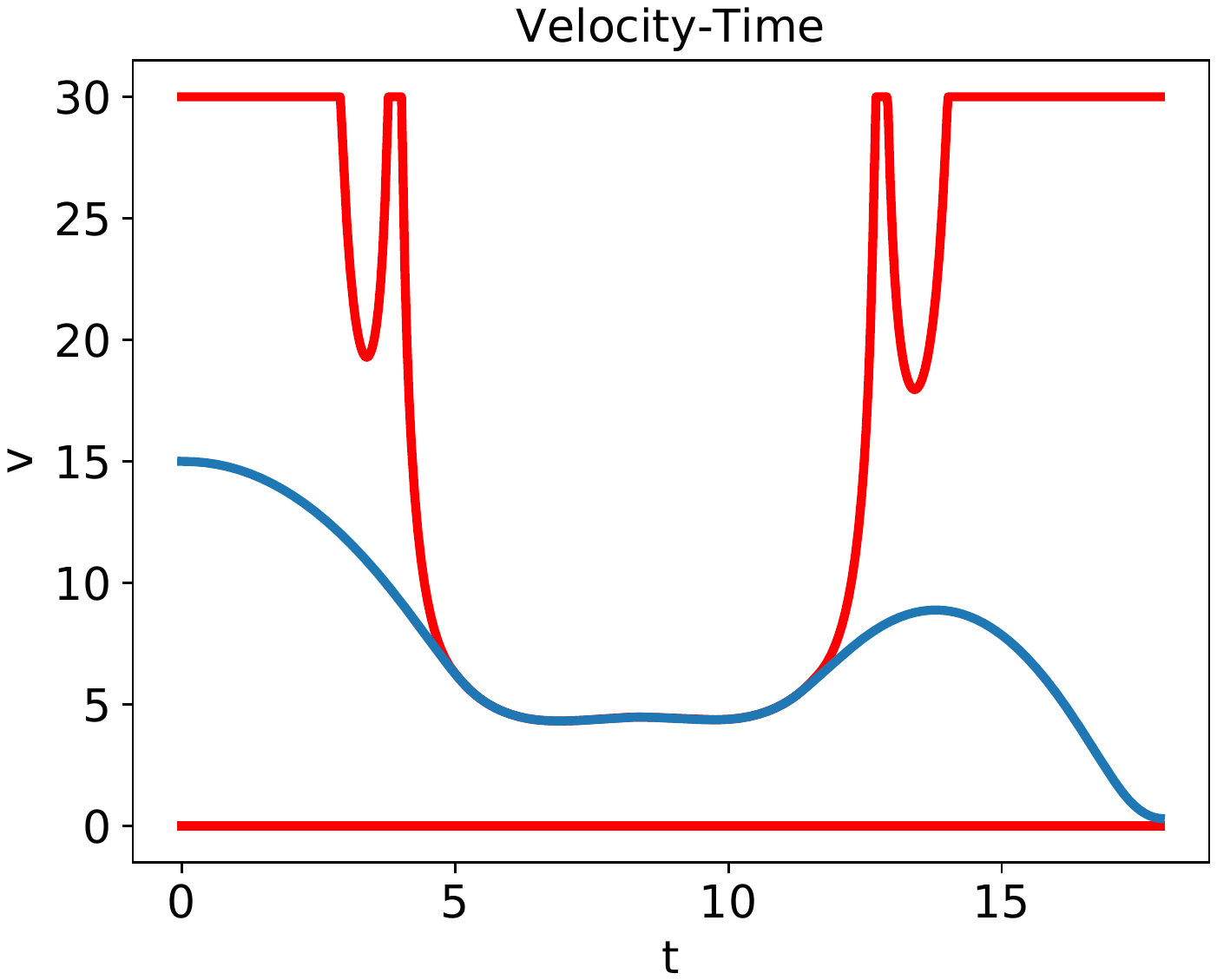} &
      \includegraphics[trim={\trside} {\tr} {\trside} {\tr}, width=\w\textwidth]{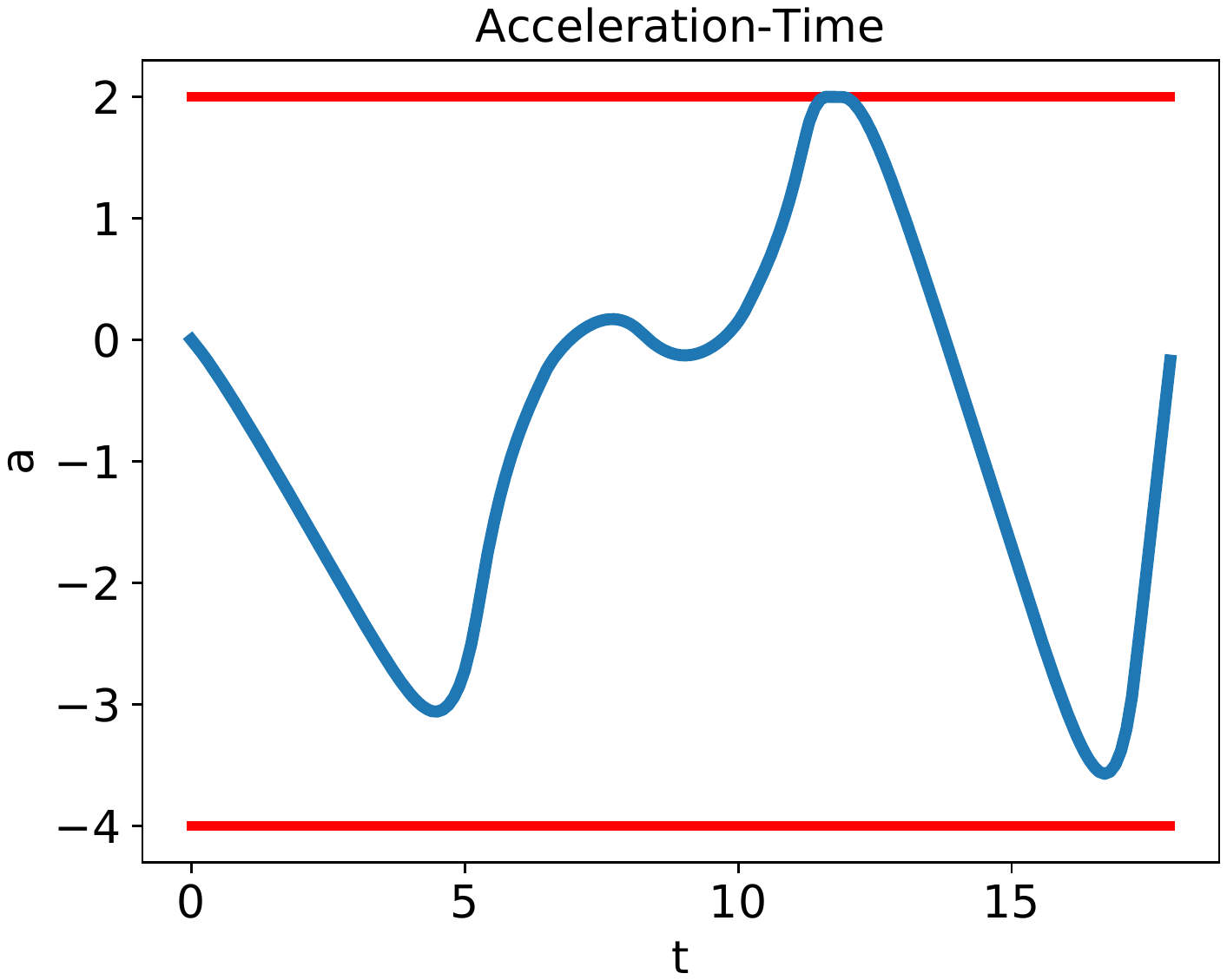} &
      \includegraphics[trim={\trside} {\tr} {\trside} {\tr}, width=\w\textwidth]{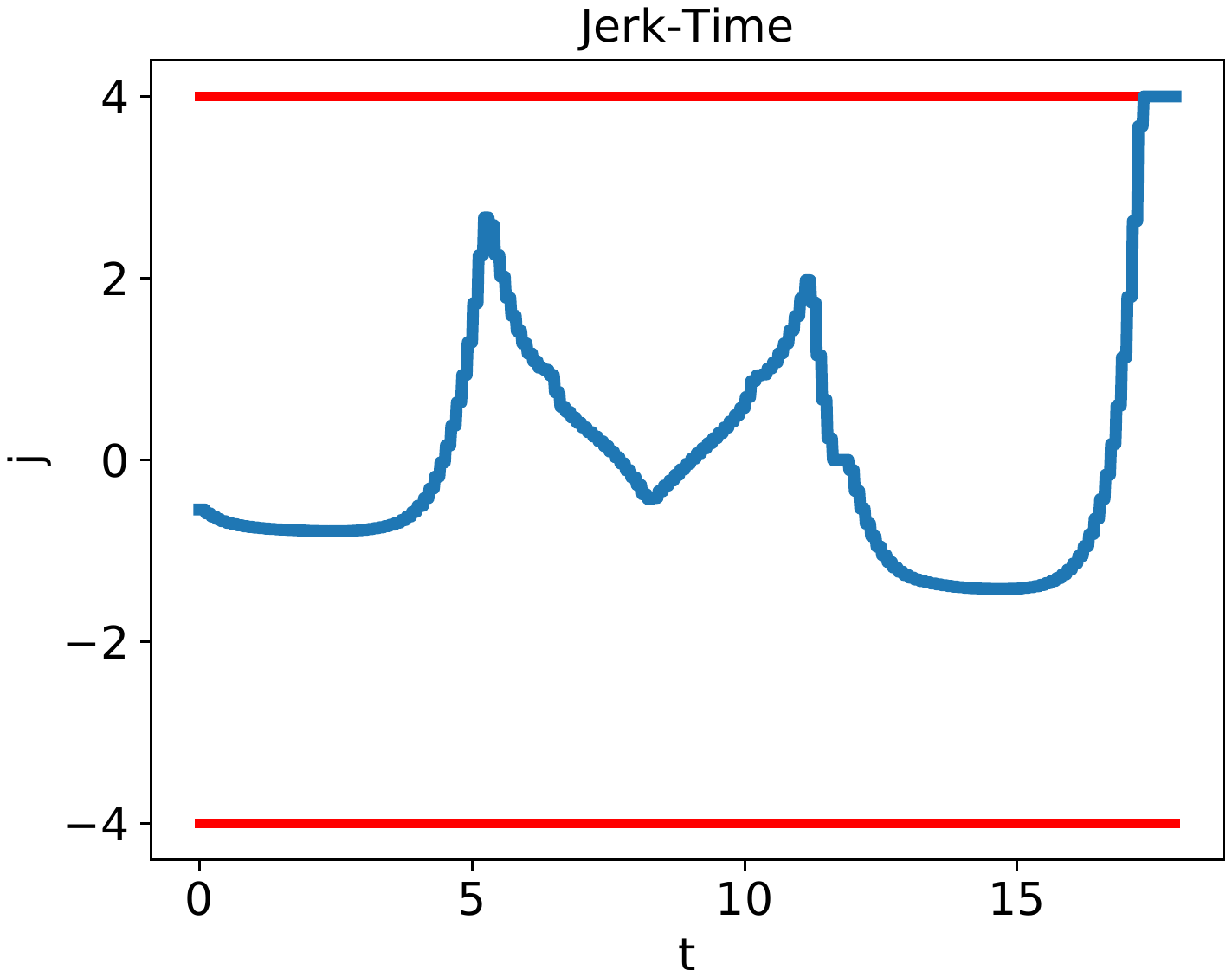} \\
      \includegraphics[trim={\trside} {\tr} {\trside} {\tr}, width=\w\textwidth]{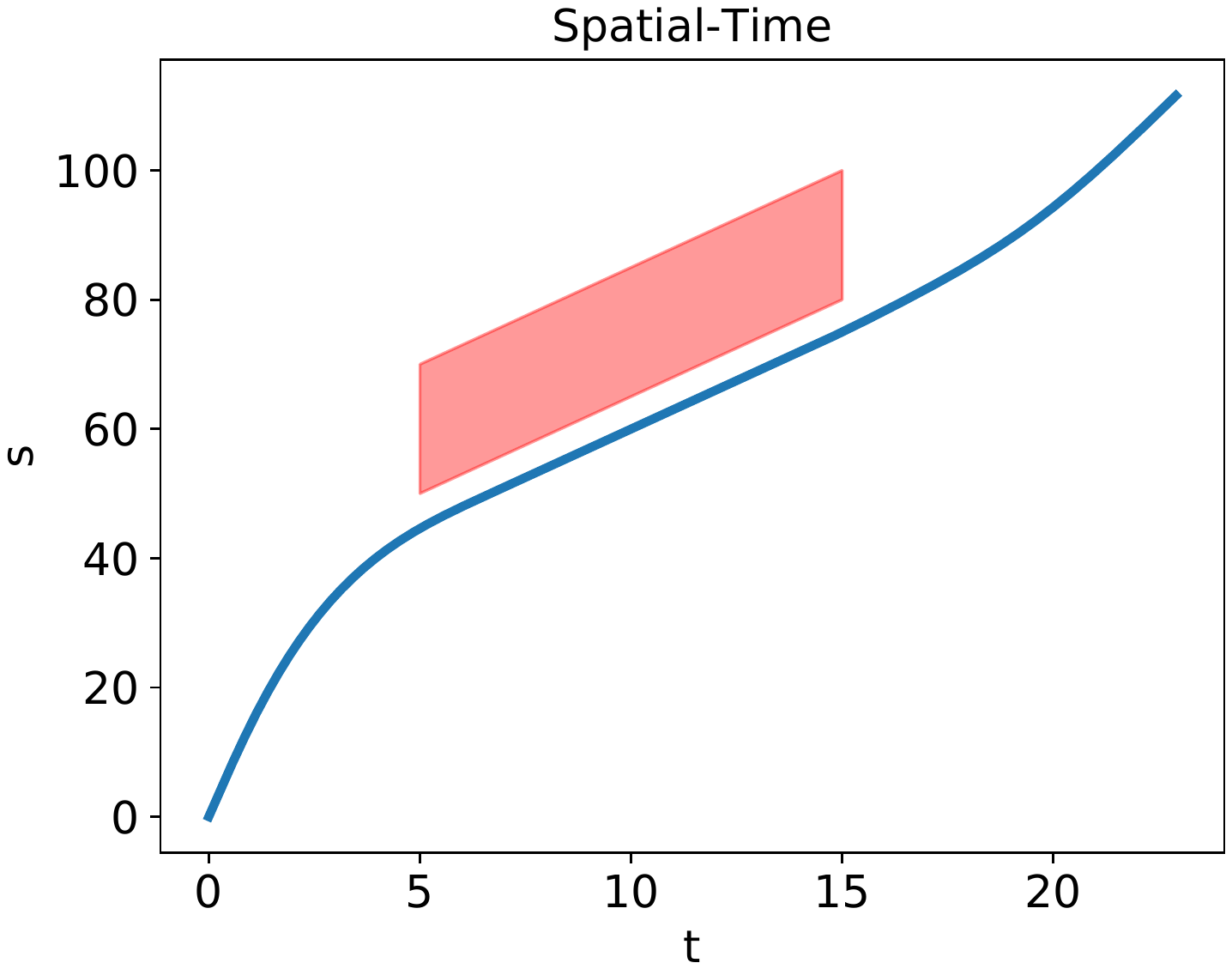} & 
      \includegraphics[trim={\trside} {\tr} {\trside} {\tr}, width=\w\textwidth]{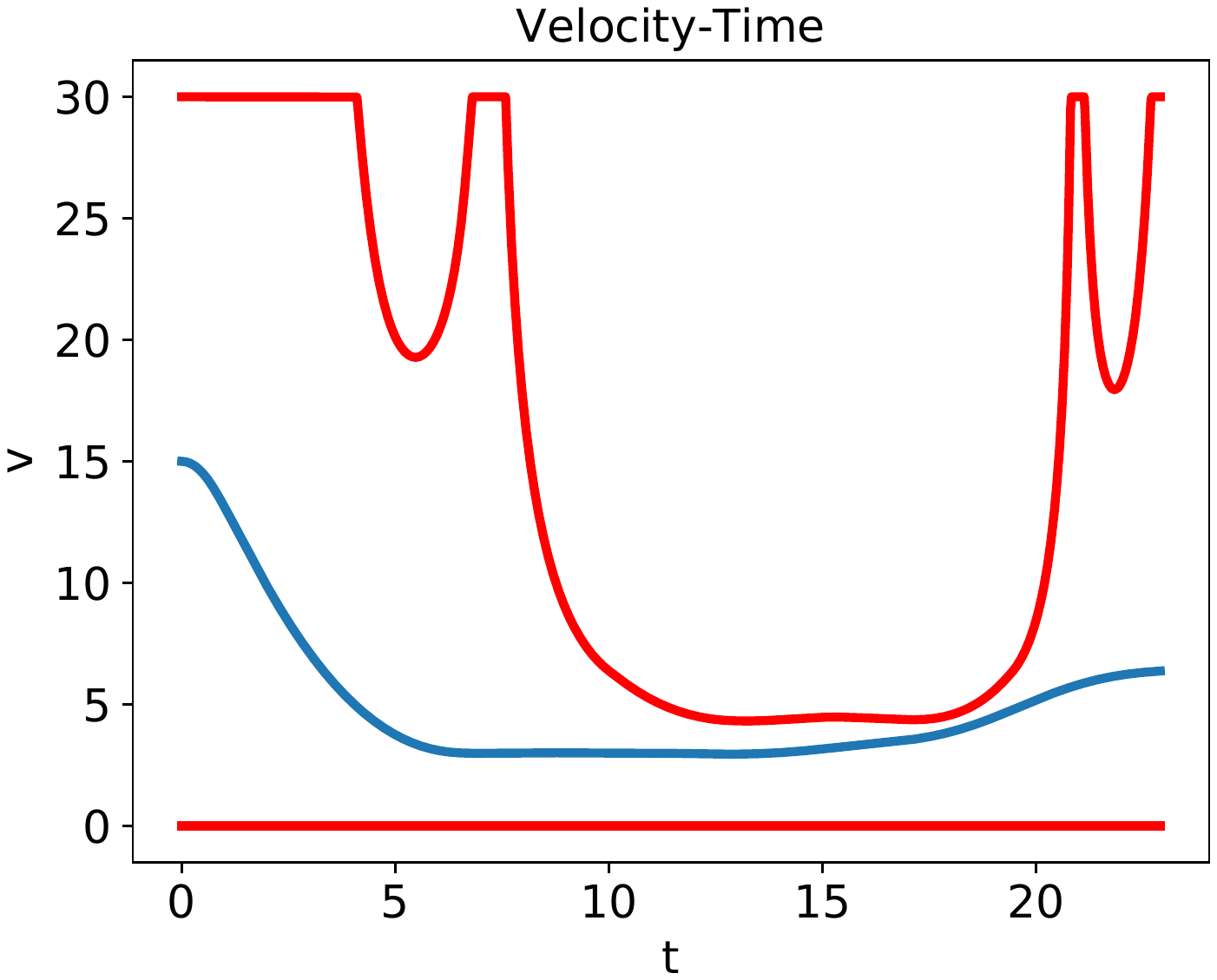} &
      \includegraphics[trim={\trside} {\tr} {\trside} {\tr}, width=\w\textwidth]{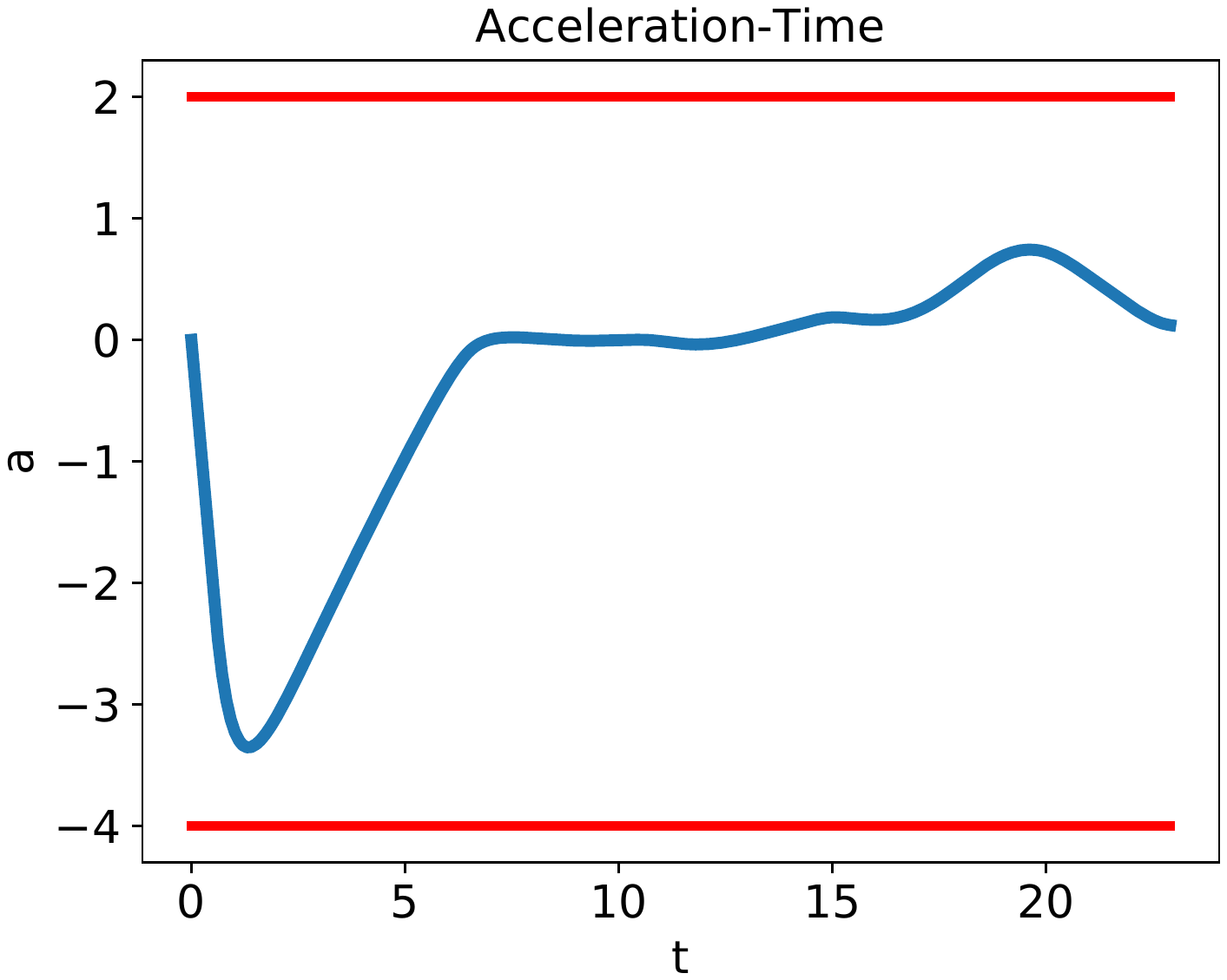} &
      \includegraphics[trim={\trside} {\tr} {\trside} {\tr}, width=\w\textwidth]{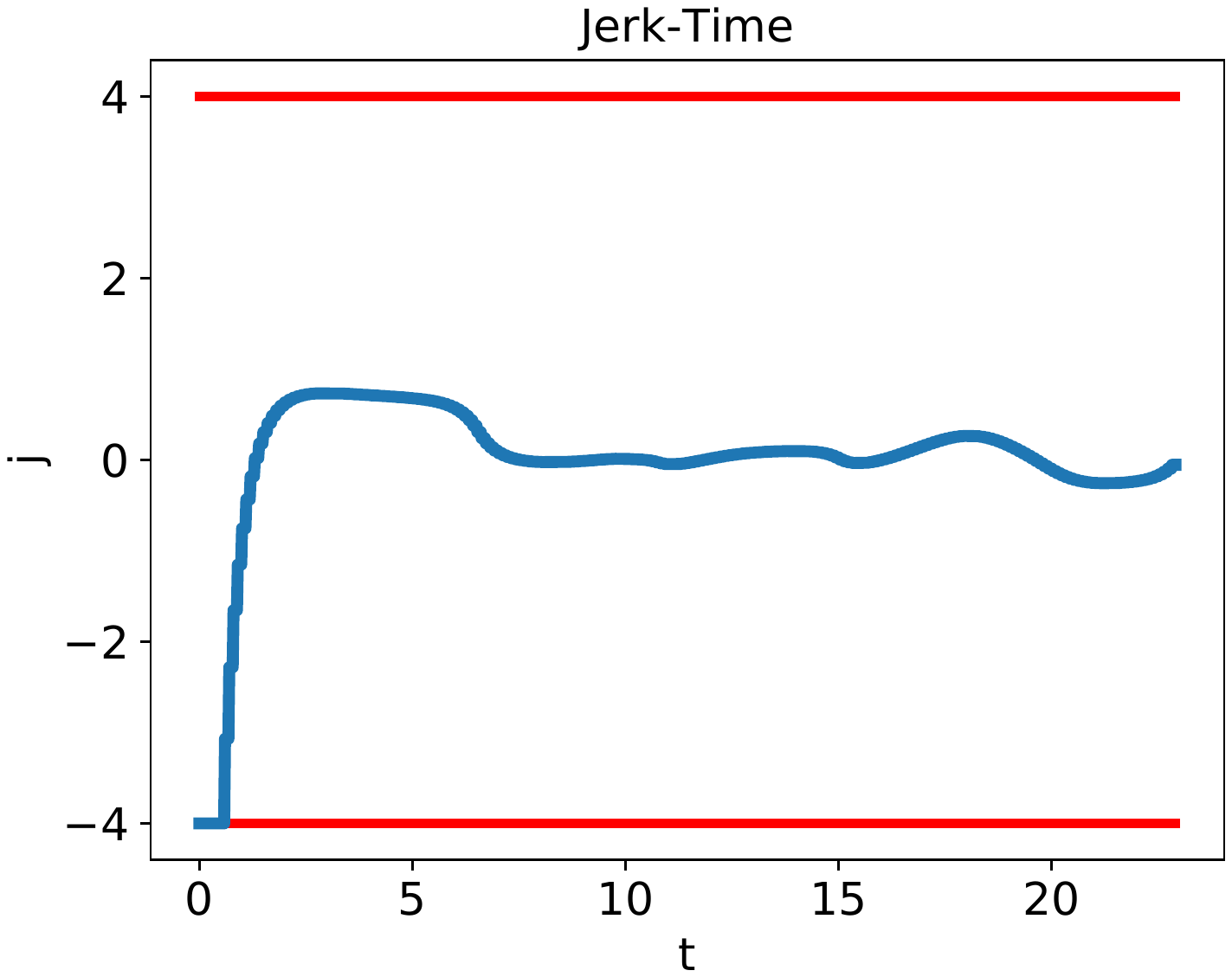} \\
\end{tabular}
\caption{Trajectory optimization result for U-turn scenario. The vehicle starts at $s_{init} = 0$, $\dot{s}_{init} = 15 m/s$, $\ddot{s}_{init} = 0 m/s^2$ with target cruise velocity 20 $m/s$. First two rows: trajectory optimization results for cruise task using two sets of parameters, one comfortable and one for sporty driving behavior. In order to show a  complete u-turn, different trajectory total time are used. The first column shows the accumulated distance versus time. The comfort trajectory has 18 $s$ time length with 121.83 $m$ spatial length, and the sporty trajectory has 15 $s$ and 128.04 $m$. The second column shows the optimized velocity (in blue) and velocity bounds (in red). The velocity lower bound is 0 $m/s$ and upper bound is computed by considering the speed limit (30 $m/s$), curvature of the guild line and maximal centripetal acceleration (2 $m/s^2$). The third column shows the optimized acceleration (in blue) and predefined acceleration bounds (in red, 2 $m/s^2$ for accelerating and -4 $m/s^2$ for braking). For sporty trajectory, the optimized acceleration hits acceleration boundaries to maximize the distance traveled while the comfortable trajectory brakes early to reduce resulted centripetal acceleration and accelerate gracefully to reach the target velocity. Third row: trajectory optimization for stop task. The vehicle starts from $s= 0$ and has a predefined full stop point at $s = 130$. The end state of the trajectory has 0 velocity and acceleration. Fourth row: trajectory optimization result for cruise task with a low speed vehicle (3 $m/s$) in the front for 10 $s$. The ego vehicles slows down and maintains a predefined safety buffer (5 $m$).}
\end{figure*}

\section{Implementation and Experiments}
The proposed two-step optimization are implemented using non-linear optimization package Interior Point OPTimizer (IPOPT) \cite{wachter2006implementation}. The source code is planned to be released as part of Baidu Apollo Open Source Platform \cite{baiduapollo}.

All experiments are carried out on an Intel Xeon 2.6GHz PC with 32GB RAM. We use a synthetic 108\textdegree U-turn scenario with turning radius roughly 10 meters ($m$), which is common for urban driving, for testing. In guide line smoothing phase, the optimization for this 17-point 150 $m$ guide line takes 948 millisecond $ms$. The input points are randomly perturbed within radius 0.05 $m$. The maximal allowed deviation is set to 0.1 $m$. Fig. 2 shows the geometrical properties of the smoothed guide line.

The testings on trajectory optimization phase are carried out on this U-turn scenario. The vehicle starts with an initial speed of 15 $m/s$ and the reference speed is set to 20 $m/s$. The following tables shows the vehicle's dynamical limits we set in the experiments:

\begin{center}
\begin{tabular}{ |c|c|c| } 
\hline
            & min.  & max. \\ 
\hline
$\dot{s}$   ($m/s$)   &0      &30        \\ 
\hline
$\ddot{s}$ ($m/s^2$)    &-4     &2         \\ 
\hline
$\dddot{s}$ ($m/s^3$)   &-4     &4         \\
\hline
$a_c$ ($m/s^2$)         &-2     &2        \\
\hline
\end{tabular}
\end{center}

The trajectory is discretized by 0.1 second ($s$). The computation time is strongly affected by the number of discretized point, the number of constraints (e.g., whether there is a preferred task state or there are obstacles exist). Our experiments show the optimization time for optimizing an 18-second trajectory ranging from 76 $ms$ to 617 $ms$ with an average of 259 $ms$. Fig. 4 shows the optimization results with different driving behaviors in cruise task, stop task and cruise task with a lead vehicle.

\section{Conclusion and Future work}
We present a new non-linear optimization based trajectory generation method for in-lane driving scenarios. This method consists of two-step optimizations with one step computing a smooth driving guide line and the next step computing a safe and comfortable trajectory along the guide line. The experiments show the algorithm can generate jerk minimal and bounded trajectory under centripetal acceleration constraint. 

For future work, we intend to expand the algorithm to tackle multi-lane driving scenarios. There are two possible directions for the expansion, both use the strategy that path planning deals with static obstacles while speed planning deals with dynamic obstacles. 

In the first direction, the guide line can be dynamically adjusted to take static obstacle avoidance into consideration. The guide line generation essentially becomes a path planning problem for static obstacle avoidance. The advantage of this direction is the rest of the algorithm can remain unchanged. However, the computation for smoothing the guide line is costly and adjustment of the guide line online may not be feasible for fast planning cycles.

Another direction is using optimization to generate the lateral trajectory/path $\boldsymbol{d}(s)$ that directly considers the projected static obstacles onto the Frenet frame. This direction is more consistent with the framework in Frenet frame planning. The benefit is the smooth guide line can be generated and stored offline and be loaded dynamically for online uses. Also, lane change actions will be handled more cleanly as lane changing can be planned by simply substituting the guide line. 

\addtolength{\textheight}{-12cm}   




\bibliography{reference}{}

\begin{thebibliography}{10}

\bibitem{baiduapollo}
Baidu apollo project.
\newblock \url{http://apollo.auto/}.

\bibitem{gulati2013nonlinear}
Shilpa Gulati, Chetan Jhurani, and Benjamin Kuipers.
\newblock A nonlinear constrained optimization framework for comfortable and
  customizable motion planning of nonholonomic mobile robots-part i.
\newblock {\em arXiv preprint arXiv:1305.5024}, 2013.

\bibitem{hwan2011anytime}
Jeong hwan Jeon, Sertac Karaman, and Emilio Frazzoli.
\newblock Anytime computation of time-optimal off-road vehicle maneuvers using
  the rrt.
\newblock In {\em Decision and Control and European Control Conference
  (CDC-ECC), 2011 50th IEEE Conference on}, pages 3276--3282. IEEE, 2011.

\bibitem{kant1986toward}
Kamal Kant and Steven~W Zucker.
\newblock Toward efficient trajectory planning: The path-velocity
  decomposition.
\newblock {\em The international journal of robotics research}, 5(3):72--89,
  1986.

\bibitem{kuwata2009real}
Yoshiaki Kuwata, Justin Teo, Gaston Fiore, Sertac Karaman, Emilio Frazzoli, and
  Jonathan~P How.
\newblock Real-time motion planning with applications to autonomous urban
  driving.
\newblock {\em IEEE Transactions on Control Systems Technology},
  17(5):1105--1118, 2009.

\bibitem{lavalle2001randomized}
Steven~M LaValle and James~J Kuffner~Jr.
\newblock Randomized kinodynamic planning.
\newblock {\em The international journal of robotics research}, 20(5):378--400,
  2001.

\bibitem{5940562}
J.~{Levinson}, J.~{Askeland}, J.~{Becker}, J.~{Dolson}, D.~{Held}, S.~{Kammel},
  J.~Z. {Kolter}, D.~{Langer}, O.~{Pink}, V.~{Pratt}, M.~{Sokolsky},
  G.~{Stanek}, D.~{Stavens}, A.~{Teichman}, M.~{Werling}, and S.~{Thrun}.
\newblock Towards fully autonomous driving: Systems and algorithms.
\newblock In {\em 2011 IEEE Intelligent Vehicles Symposium (IV)}, pages
  163--168, June 2011.

\bibitem{miller2008team}
Isaac Miller, Mark Campbell, Dan Huttenlocher, Frank-Robert Kline, Aaron
  Nathan, Sergei Lupashin, Jason Catlin, Brian Schimpf, Pete Moran, Noah Zych,
  et~al.
\newblock Team cornell's skynet: Robust perception and planning in an urban
  environment.
\newblock {\em Journal of Field Robotics}, 25(8):493--527, 2008.

\bibitem{schlechtriemen2016wiggling}
Julian Schlechtriemen, Kim~Peter Wabersich, and Klaus-Dieter Kuhnert.
\newblock Wiggling through complex traffic: Planning trajectories constrained
  by predictions.
\newblock In {\em Intelligent Vehicles Symposium (IV), 2016 IEEE}, pages
  1293--1300. IEEE, 2016.

\bibitem{urmson2008autonomous}
Chris Urmson, Joshua Anhalt, Drew Bagnell, Christopher Baker, Robert Bittner,
  MN~Clark, John Dolan, Dave Duggins, Tugrul Galatali, Chris Geyer, et~al.
\newblock Autonomous driving in urban environments: Boss and the urban
  challenge.
\newblock {\em Journal of Field Robotics}, 25(8):425--466, 2008.

\bibitem{wachter2006implementation}
Andreas W{\"a}chter and Lorenz~T Biegler.
\newblock On the implementation of an interior-point filter line-search
  algorithm for large-scale nonlinear programming.
\newblock {\em Mathematical programming}, 106(1):25--57, 2006.

\bibitem{werling2010optimal}
Moritz Werling, Julius Ziegler, S{\"o}ren Kammel, and Sebastian Thrun.
\newblock Optimal trajectory generation for dynamic street scenarios in a
  frenet frame.
\newblock In {\em Robotics and Automation (ICRA), 2010 IEEE International
  Conference on}, pages 987--993. IEEE, 2010.

\bibitem{ziegler2014making}
Julius Ziegler, Philipp Bender, Markus Schreiber, Henning Lategahn, Tobias
  Strauss, Christoph Stiller, Thao Dang, Uwe Franke, Nils Appenrodt,
  Christoph~Gustav Keller, et~al.
\newblock Making bertha drive-an autonomous journey on a historic route.
\newblock {\em IEEE Intell. Transport. Syst. Mag.}, 6(2):8--20, 2014.

\end{thebibliography}
\bibliographystyle{plain}

\end{document}